\documentclass{article}
\usepackage{spconf,amsmath,epsfig}
\usepackage{url}
\usepackage{amssymb}
\usepackage{enumitem}
\usepackage{booktabs}
\usepackage{subcaption}
\usepackage{wrapfig}
\usepackage{float}

\let\OLDthebibliography\thebibliography
\renewcommand\thebibliography[1]{
  \OLDthebibliography{#1}
  \setlength{\parskip}{0pt}
  \setlength{\itemsep}{0pt plus 0.3ex}
}

\begin{document}\sloppy

\def\x{{\mathbf x}}
\def\L{{\cal L}}

\title{Manga109Dialog: A Large-scale Dialogue Dataset\\ for Comics Speaker Detection}
%
\name{Yingxuan Li, Kiyoharu Aizawa, Yusuke Matsui}
\address{The University of Tokyo\\
\small\url{{li, aizawa, matsui}@hal.t.u-tokyo.ac.jp}
}

\maketitle

\begin{abstract}
The expanding market for e-comics has driven the development of automated methods for analyzing comics. To enhance the machine's understanding of comics, an automated method is essential for linking text in comics to characters that speak those words.
In this study, we developed Manga109Dialog\footnote{Dataset and code are available at \url{https://github.com/liyingxuan1012/Manga109Dialog}.}, which is the world's largest speaker-to-text annotation dataset for comics, containing 132,692 pairs. 
We proposed a novel deep learning-based method using scene graph generation models. To tailor the unique features of comics, we enhanced the performance by considering the frame reading order. 
Our experiments with Manga109Dialog show that our scene-graph-based approach outperforms existing methods, achieving a prediction accuracy of over 75\%, thus establishing a robust benchmark for speaker detection in comics.

\end{abstract}
\begin{keywords}
Dataset, Comic, Scene graph generation
\end{keywords}

\section{Introduction}
\label{sec:intro}

The market for e-comics has been expanding rapidly with the popularization of digital devices. In 2022, e-comics made up 66.2\% of the Japanese comics market~\cite{shuppankagaku2023}.
The development of e-comics has led to increased interest in automated methods to analyses comics.
To support these techniques, reliable large-scale dialogue datasets are required for improved computational understanding of comics. Additionally, automated methods to detect characters to whom text is attributed are necessary for effective speaker detection.

\begin{figure}[t]
    \centering
    \includegraphics[width=0.8\linewidth]{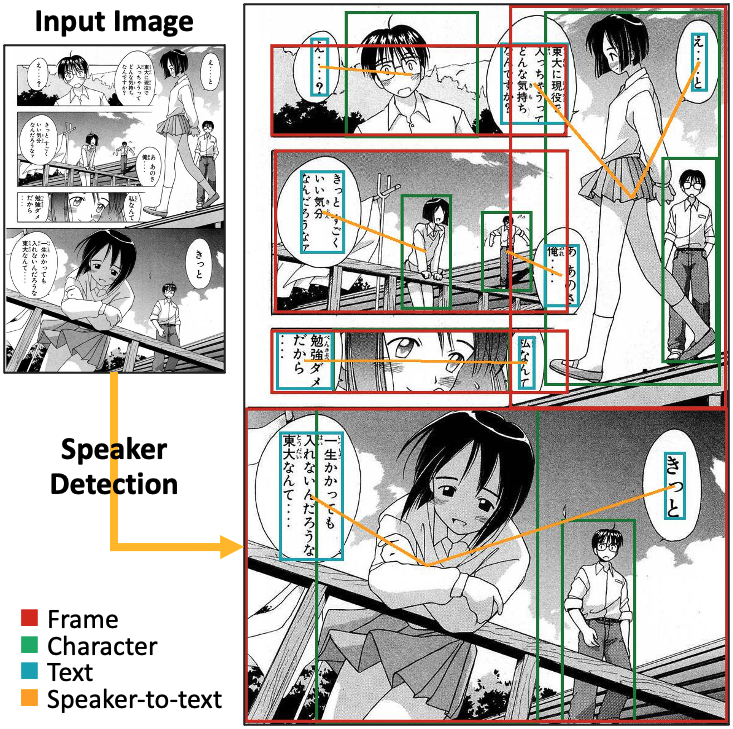}
    \caption{An example of speaker detection. Courtesy of Akamatsu Ken.}
    \label{fig:intro}
\end{figure}

An example of speaker detection is shown in Fig.~\ref{fig:intro}. Given an image from comics, the system can detect characters, text, and frame regions, and can then automatically predict the character to which the text is attributed.
This research can be applied to various tasks, such as automatic character assignment for text-to-speech reading, automatic translation according to characters' personalities, inference of character relationships and stories.

To address the significant gap in reliable speaker-to-text annotation datasets for comics, we developed Manga109Dialog. To the best of our knowledge, this is the largest comics dialogue dataset ever created.
In order to validate the reliability and accuracy of Manga109Dialog, we conducted comprehensive speaker detection experiments on it.
Since previous approaches were primarily based on the rule that the character nearest to a given text is most likely to be the speaker~\cite{rigaud2015speech,abe2020}, they may have difficulty making correct predictions in some complex cases.
As an example, consider the text at bottom right in Fig.~\ref{fig:intro}. If we chose the character nearest to the text as the speaker, the boy would be incorrectly predicted as the speaker rather than the girl.
Therefore, the relationship between characters and text should be considered to make a correct prediction.

To address this challenge, we propose a novel deep learning-based approach. We leverage scene graphs, a powerful tool for describing visual relationships. 
The task of scene graph generation (SGG) involves detecting objects and their relationships within an image. 
Given the successful application of SGG models in analyzing real-world datasets, such as the Visual Genome dataset~\cite{krishna2017visual}, we propose their adaptation for comics speaker detection. 
In addition to employing the standard framework of SGG models, we incorporate frame information, as we find that the speaker tends to appear in the same frame as the text.

Traditional metrics used to evaluate SGG lack suitability because the amount of speaker-to-text pairs varies from page to page.
Therefore, we present an evaluation metric well suited for this specific task.

The contributions are summarized as follows.
\begin{itemize}[leftmargin=*]
 \setlength{\itemsep}{-0.1ex}
 \item We constructed Manga109Dialog, an annotation dataset of associations between speakers and texts. This is the largest comics dialogue dataset in the world.
 \item We proposed a deep learning-based approach for comics speaker detection using SGG models. We enhance the results by introducing frame information in the relationship prediction stage.
 \item We established a new benchmark for evaluation, setting a standard for future research in this domain.
\end{itemize}

Experimental results demonstrated that Manga109Dialog provides a challenging yet realistic benchmark for comics speaker detection.
Moreover, our proposed approach outperformed conventional rule-based methods by 5\%, highlighting its effectiveness in complex cases.

\section{Related Work}

\subsection{Comics Speaker Dataset}
Given the scarcity of available comics speaker datasets, Abe et al.~\cite{abe2020} first constructed a dataset based on Manga109~\cite{aizawa2020building}, which contains 109 Japanese comics and annotations for both character and text regions. 
Although Abe's work is pioneering, the dataset has some notable limitations.
First, it does not have explicit links between the text and the character in the comics -- it only adds the character name to the text without specifying the bounding box of the character, which poses a challenge when the same character appears multiple times on a page. 
Additionally, the tendency to annotate simpler relationships results in exclusion of more complex cases. 
To address these shortcomings, we developed Manga109Dialog, employing a unique annotation standard to provide a more comprehensive and applicable dataset.

\subsection{Comics Speaker Detection}
Traditional approaches in comics speaker prediction typically focus on predicting the relationships between characters and text without detecting specific object regions. 
These methods, primarily rule-based~\cite{rigaud2015speech,abe2020}, determine the speaker by calculating the distances between text and characters, which often makes them struggle in complex cases.
Recently, there has been a shift towards deep learning-based approaches, such as extracting features from objects or employing time-series learning techniques~\cite{yamamoto2018,nakagawa2019}.
Although they demonstrated the potential of deep learning-based approaches, they were limited by using only partial datasets and showed suboptimal performance.

This gap underscores the need for a complete dataset and a unified standard for evaluation. 
In our study, we introduce object detection into the speaker prediction evaluation, aligning more closely with real-world applications. 
We also develop a comprehensive benchmark for this task.

\subsection{Scene Graph Generation (SGG)}
Scene graph is a data structure for describing objects in a scene.
Since scene graphs provide rich semantic features, they can distinguish images or videos more accurately and describe them more precisely.
Scene graphs have demonstrated success in various visual tasks, including image retrieval, image captioning, and visual question answering~\cite{johnson2015image,yao2018exploring,tang2019learning}.

\section{Manga109Dialog}
\label{sec:dataset}


\begin{figure}[t]
    \centering
    \includegraphics[width=0.75\linewidth]{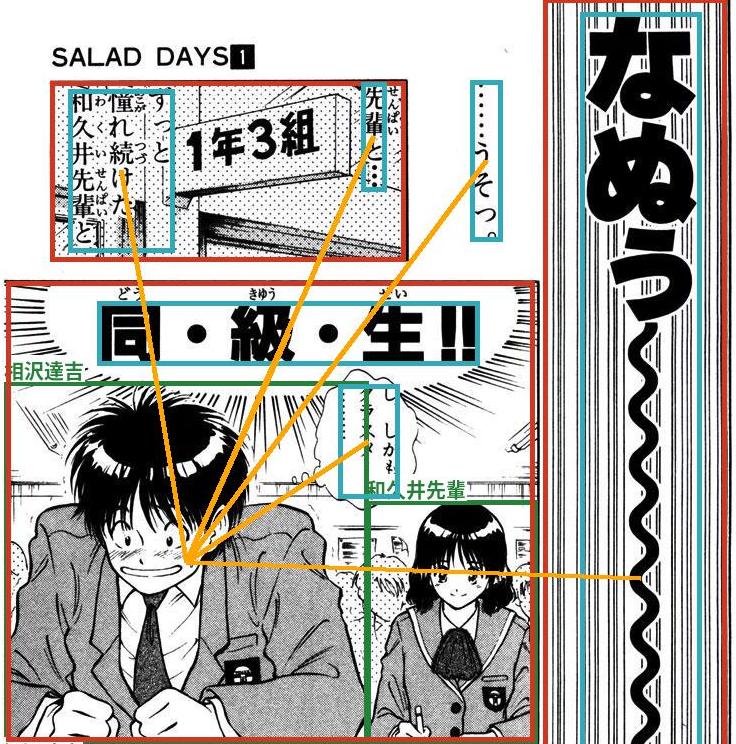}
    \caption{Annotation rules. Courtesy of Inokuma Shinobu.}
    \label{fig:rule}
\end{figure}

\subsection{Dataset Construction}
In this section, we show the details of how we constructed Manga109Dialog.
Aiming to produce a more complete and applicable dataset for speaker detection in comics, we created our dataset in a distinct annotation standard. 
Instead of connecting the character name to the text, we connect the bounding box of the speaker to the text.
The annotations were mainly based on the following rules.
\begin{itemize}[leftmargin=*]
 \setlength{\itemsep}{-0.1ex}
 \item If the character to whom the text is attributed appears only one on the page, link the bounding box of the character to the text regardless of the position.
 \item If more than one character is on the page, the bounding box of the speaker in the same frame as the text is linked in priority. If the speaker is not in the same frame, it is determined by the reading order (from top to bottom, right to left), as shown in Fig.~\ref{fig:rule}.
 \item When there is more than one speaker, we link the text to all speakers.
 \item Texts spoken by ``Others'' are annotated if we can specify the bounding box of the speaker, and are excluded otherwise.
\end{itemize}

More details and examples of annotation rules are shown in the supplementary material.

\begin{table}[t]
    \centering
    \caption{An overview of Manga109Dialog.}
    \begin{tabular}{@{}ll@{}}
        \toprule
        Item & Number of elements \\ 
        \midrule
        Images & 10,602 \\
        Annotated Images & 9,904 \\
        Texts & 147,887 \\
        Speaker-to-Text Pairs & \\
        ~~~~\textit{Easy} & 111,959 \\
        ~~~~\textit{Hard} & 20,733 \\
        ~~~~\textit{Total} & 132,692 \\
        Pairs / Page (\textit{Total}) & 6.70 \\
        \bottomrule
    \end{tabular}
    \label{tab:statistics}
\end{table}

\subsection{Data Analysis}
\label{sec:data_analysis}
Creating the annotations for Manga109Dialog took approximately three months. An overview of the dataset is presented in Table~\ref{tab:statistics}. 
Manga109Dialog comprises 9,904 images featuring speaker-to-text pairs, averaging 6.70 pairs per page.

\begin{figure}[t]
    \centering
    \begin{minipage}[c]{0.32\linewidth}
        \centering
        \includegraphics[width=\linewidth]{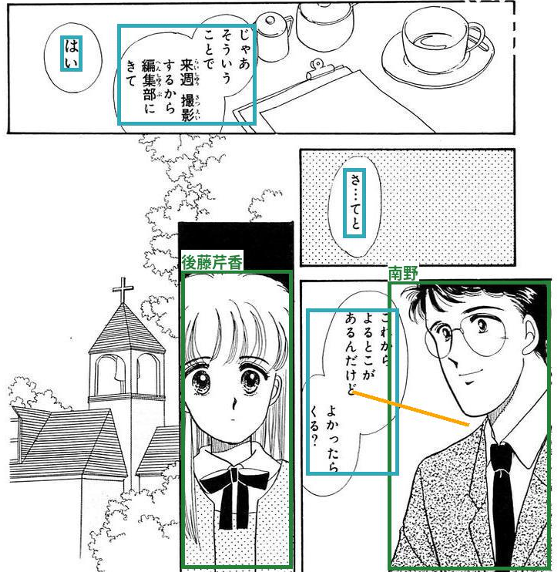}
        \subcaption{\textit{Easy}.}
    \end{minipage}
    \begin{minipage}[c]{0.32\linewidth}
        \centering
        \includegraphics[width=\linewidth]{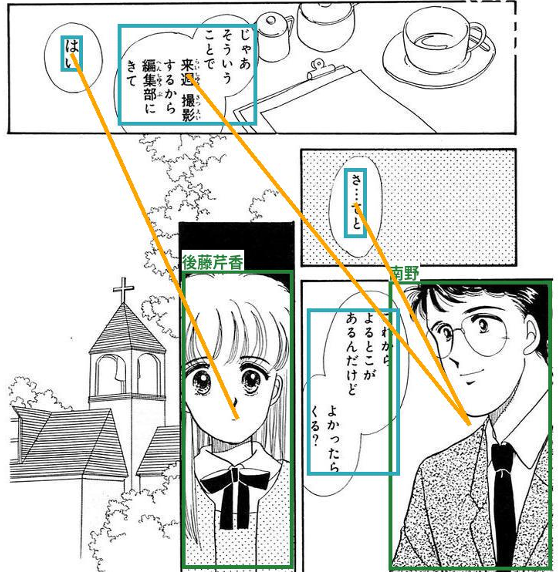}
        \subcaption{\textit{Hard}.}
    \end{minipage}
    \begin{minipage}[c]{0.32\linewidth}
        \centering
        \includegraphics[width=\linewidth]{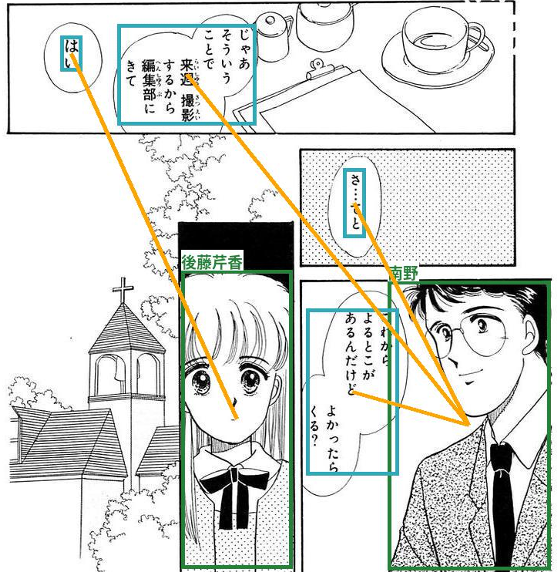}
        \subcaption{\textit{Total}.}
    \end{minipage}
    \caption{An example of annotations on three prediction difficulties. Courtesy of Kurita Riku.}
    \label{fig:dataset_div}
\end{figure}

We further divided our annotations into \textit{Easy} and \textit{Hard} by prediction difficulties.
If the speaker is in the same frame as the text, the text is considered \textit{Easy}; otherwise, it is considered \textit{Hard}.
\textit{Total} contains all pairs of speakers and texts.
An example of annotations of varying difficulties is given in Fig.~\ref{fig:dataset_div}.
We conducted experiments under these three difficulties to evaluate speaker detection methods more appropriately.

\section{Approach}
\subsection{Problem Setting}
Before introducing our proposed scene-graph-based approach, let us first define our problem setting. 
Unlike existing studies on comics speaker prediction, we incorporate the steps for detecting and identifying object regions. 

Given an input image $I$, our task is twofold.
First, we localize character regions and text regions.
Then, for all combinations of characters and texts, we calculate the relationship score of each combination. The character with a higher score is more likely to be the speaker.

\begin{figure*}[t]
    \centering
    \includegraphics[width=0.85\linewidth]{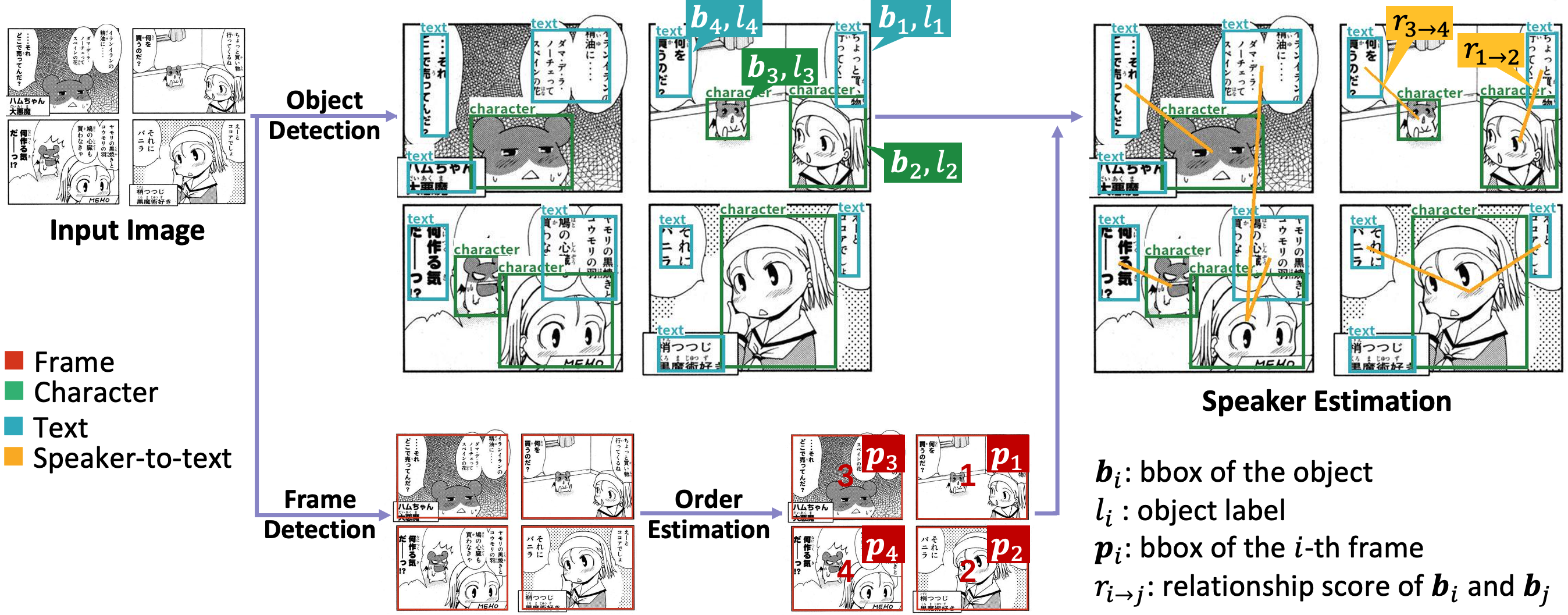}
    \caption{Framework of the proposed method. Courtesy of Arai Satoshi.}
    \label{fig:pipeine}
\end{figure*}

An example of the pipeline is shown in Fig.~\ref{fig:pipeine}. We represent each region as a bounding box $\mathbf{b} = [x, y, w, h]^\top \in \mathbb{R}^4$.
For each region, we predict an object label $l \in \{\mathtt{character}, \mathtt{text},\mathtt{background}\}$.
For the combination of character $i$ and text $j$, the relationship score between $\mathbf{b}_i$ and $\mathbf{b}_j$ is represented as $r_{i \to j} \in \mathbb{R}$.
This problem setting is equivalent to a standard SGG task, where the number of object classes is two (\textit{character} or \textit{text}) and the number of relationship labels is one (\textit{speak}).
In addition, we develop a frame detector and a frame order estimator to detect frame bounding boxes ($\mathbf{p} = [x, y, w, h]^\top \in \mathbb{R}^4$) and determine their reading order. We later use this information to enhance our system.

\subsection{Speaker Detection via SGG}
\label{sec:sgg}
Our approach is based on the MOTIFS pipeline~\cite{zellers2018neural,tang2020unbiased}, the most representative SGG model. Its process is divided into three stages.
The first stage is object detection. Given an image $I$, an object detector $f_\mathrm{detect}$ outputs a set of $N$ tuples.
\begin{equation}
\{ (\mathbf{b}_i, \mathbf{f}_i, \mathbf{l}_i) \}_{i=1}^N = f_\mathrm{detect}(I).
\label{eq:detect}
\end{equation}
Each tuple consists of a bounding box $\mathbf{b}_i \in \mathbb{R}^4$, a feature vector $\mathbf{f}_i \in \mathbb{R}^{4096}$, and a vector of class label probabilities $\mathbf{l}_i \in \mathbb{R}^3$.
Unlike MOTIFS, we employ a Faster-RCNN~\cite{ren2015faster} with a ResNeXt-101-FPN~\cite{lin2017feature,xie2017aggregated} backbone for this step.

Next, we fuse the detected feature vectors to produce a richer representation. 
We feed the set of the feature vectors and label probability vectors $\{(\mathbf{f}_i, \mathbf{l}_i)\}_{i=1}^N$ into the fusion module $f_\mathrm{fuse}$ to output enhanced features $\{\mathbf{d}_i\}_{i=1}^N$.
\begin{equation}
\{ \mathbf{d}_i \}_{i=1}^N = f_\mathrm{fuse}(\{ (\mathbf{f}_i, \mathbf{l}_i) \}_{i=1}^N).
\label{eq:fuse}
\end{equation}
Here, $\mathbf{d}_i \in \mathbb{R}^{512}$ is an enhanced representation for $\mathbf{b}_i$.
We adopt bidirectional LSTM models~\cite{hochreiter1997long} for this step. An additional LSTM is used to predict the object label $l_i$ from $\mathbf{l}_i$.

In the last stage, we calculate the relationship score for each pair among the $N$ bounding boxes, resulting in \(N^2\) possible combinations. For example, the score of $\mathbf{b}_i$ and $\mathbf{b}_j$ ($r_{i \to j}\in \mathbb{R}$) is computed as
\begin{equation}
r_{i \to j} = w(i, j) g(\mathbf{d}_i, \mathbf{d}_j, \mathbf{f}_{i, j}).
\label{eq:relation}
\end{equation}
\begin{equation}
\mathbf{f}_{i, j}=f_\mathrm{extract}( \mathbf{b}_i, \mathbf{b}_j, \mathbf{f}_i, \mathbf{f}_j ).
\label{eq:iou_feature}
\end{equation}
The function $g$ inputs the union box's RoI feature $\textbf{f}_{i, j}$ and enhanced features $\mathbf{d}_i, \mathbf{d}_j$ and outputs a score.
Here, $g$ consists of simple learnable matrices and a Softmax predicate classifier. 
The feature vector $\textbf{f}_{i, j}$ is extracted from bounding boxes $\mathbf{b}_i, \mathbf{b}_j$ and their feature vectors $\mathbf{f}_i, \mathbf{f}_j$.
The weight function $w(i, j) \in \mathbb{R}$ can be any function. If $w(i, j)=1$, the entire pipeline is the same as that of MOTIFS.

\subsection{Use of Reading Order}
Due to the unique feature of comics that the speaker is more likely to appear in the same or next frame that the text belongs to, we add frame information to the SGG model.

We train another Faster-RCNN model to detect the frame regions from the input image.
To estimate the reading order of these frames, we follow the algorithm described in Kovanen and Ikuta's study~\cite{kovanen2015layered,ikuta2023}.
As shown in Fig.~\ref{fig:pipeine}, the outcome is a series of ordered frame bounding boxes $\mathbf{p}_1, \mathbf{p}_2, \dots$, where $\mathbf{p}_i \in \mathbb{R}^4$ is a bounding box of the $i$-th frame.
We show the details of this process and results of frame detection and reading order estimation in the supplementary material.

For each object, we calculate to which frame it belongs and obtain the reading order.
We empirically found that the model performed best when the weight function $w(i, j)$ was in the form of Eq.~\ref{eq:weight}:
\begin{equation}
w(i, j) = \frac{1}{2 + |k_i - k_j|}.
\label{eq:weight}
\end{equation}
Here, $k_i$ represents the reading order of the frame $\mathbf{p}$ to which $\mathbf{b}_i$ belongs. 
The closer the reading orders of $\mathbf{b}_i$, $\mathbf{b}_j$ are, the higher the score of $w(i, j)$.
Therefore, we can enhance prediction accuracy by incorporating frame orders and adjusting relationship scores accordingly. 


\subsection{Evaluation Metrics}
\label{sec:evaluation}
\begin{figure}[t]
  \centering
  \includegraphics[width=\linewidth]{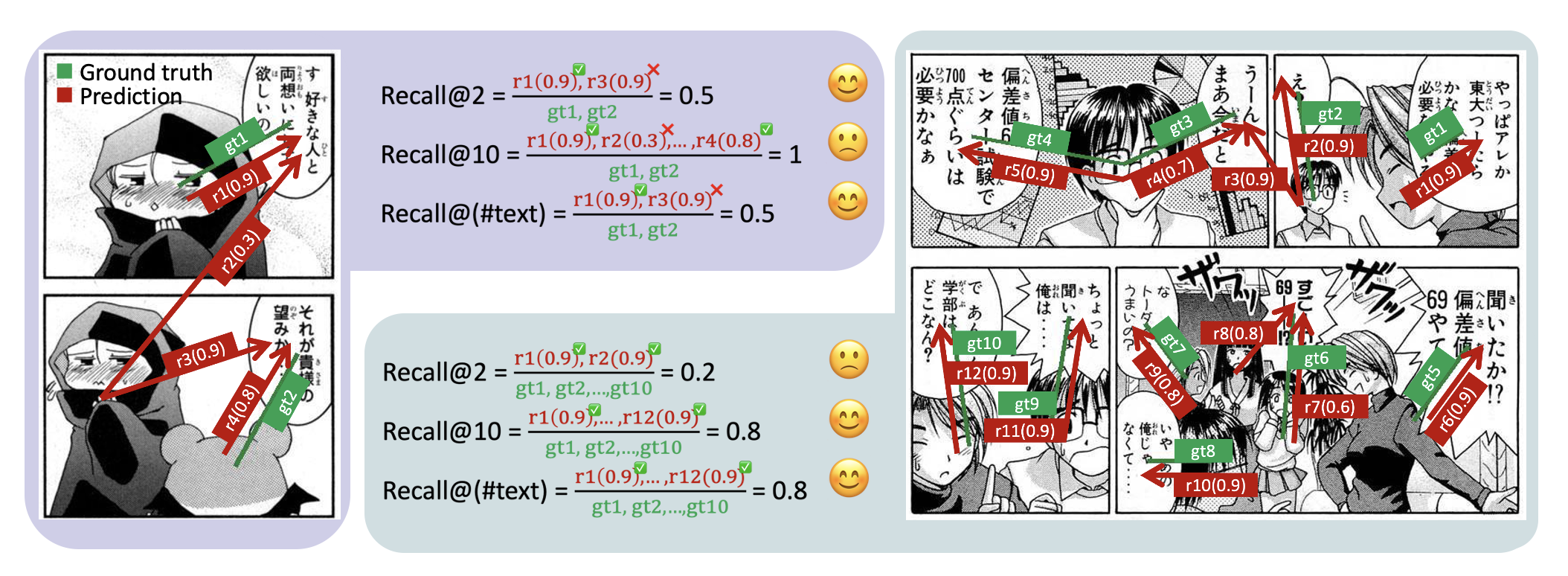}
  \caption{Comparison of using Recall@K and Recall@(\#text). Courtesy of Arai Satoshi, Akamatsu Ken.}
  \label{fig:recall}
\end{figure}

The earliest and the most widely accepted evaluation metric for SGG models is Recall@K, traditionally predefined at 20, 50, or 100~\cite{lu2016visual}.
However, since the number of texts in comics fluctuates significantly from image to image, using a fixed $K$ value does not provide a fair comparison.
Take Fig.~\ref{fig:recall} as an example.
When a large $K$ is fixed, Recall@K can cover almost all major combinations of objects if the number of texts is small, which makes the accuracy very close to 100\% (left side of Fig.~\ref{fig:recall}). 
When a small $K$ is fixed, the score is always low if there are many texts (right side in Fig.~\ref{fig:recall}).

To establish a more reliable benchmark for the task of comics speaker detection, we introduce a new evaluation metric called Recall@(\#text), which measures the Recall of selected predictions covering all texts on a page.
For each text, we choose the subject-predicate-object triplet containing it with the highest score as the prediction. 
In cases where a single text has $N$ speakers, we select the top $N$ triplets. 
We consider a triplet correct if the object regions are detected with an IoU greater than 0.65 and correctly labeled.

\section{Experiments}

\subsection{Experimental Setup}
\label{sec:setting}
\textbf{Baseline:}
The simplest rule-based method is \textbf{``shortest distance''}, where we assume the text is spoken by the closest character.
A slightly modified version is described as \textbf{``frame distance''}. We prioritize characters in the same frame as the text. If there are no characters in the frame, we make predictions using the same rule as ``shortest distance''.
Besides, we propose a deep learning-based baseline, where the relationship score was calculated only by feeding the union feature $\textbf{f}_{i, j}$ obtained from Eq.~\ref{eq:iou_feature} into a fully connected layer.

\vspace{5pt}
\noindent\textbf{Datasets:}
We divided Manga109Dialog into training and test sets with ratios of 70\% and 30\%, respectively.
Furthermore, to see our model's behavior on different styles of comics or languages, we tested it on eBDtheque, which comprises 100 images from French, American, and Japanese comics~\cite{guerin2013ebdtheque}. 

\vspace{5pt}
\noindent\textbf{Evaluation:}
In the task of SGG, the performance of the models can be evaluated in three protocols.
(1) \textbf{Predicate Classification (PredCls)}: predicting the relationship between two objects, given an image $I$, object bounding boxes $\{\mathbf{b}_i \}$, and object labels $\{l_i\}$. 
(2) \textbf{Scene Graph Classification (SGCls)}: predicting object labels and relationships, given an image and bounding boxes. 
(3) \textbf{Scene Graph Detection (SGDet)}: detecting object bounding boxes and predicting their labels and relationships using only the given image.
We executed the model for these three tasks to evaluate our proposed method.

\vspace{5pt}
\noindent\textbf{Implementation Details:}
Because object detection is the most time-consuming step, we pre-trained a Faster-RCNN model as our object detector.
Additionally, we pre-trained another Faster-RCNN as the frame detector.
The Mean Average Precision (mAP) of the two detectors reached 86.09\% and 96.48\%, respectively.
We present the details of them in the supplementary material.
In the stage of SGG, we trained our model on a single NVIDIA A100 GPU, using the SGD optimizaer with a batch size of 4. 
We used the WarmupReduceLROnPlateau as the learning scheduler and set the initial learning rate to $4 \times 10^{-2}$. 
The loss was the sum of the cross entropy for objects and relationships.

\begin{table}[t]
    \centering
    \caption{Recall@(\#text) for PredCls.}
    \begin{tabular}{@{}llll@{}} \toprule
              & \multicolumn{3}{c}{Manga109Dialog} \\ \cmidrule(l){2-4}
        Method & \textit{Easy} & \textit{Hard} & \textit{Total} \\ \midrule
        Rule-based               & & & \\
        ~~~~Short distance       & 71.43 & 22.72 & 63.41 \\
        ~~~~Frame distance       & 81.55 & 22.06 & 71.53 \\
        Deep learning-based             & & & \\
        ~~~~Only w/ union feature   & 83.20 & 28.50 & 73.99 \\
        ~~~~Proposed w/o frame   & 84.73 & 29.55 & 75.48 \\   
        ~~~~Proposed w/ frame    & \textbf{84.77} & \textbf{30.65} & \textbf{75.69} \\ \bottomrule   
    \end{tabular}
    \label{tab:predcls}
\end{table}

\subsection{Quantitative Results}
\label{sec:results}
We ran our model under three tasks (PredCls, SGCls, SGDet).
Because our study focuses on speaker prediction, we only show the results under PredCls in Table~\ref{tab:predcls}, meaning that the model only needs to find speaker-to-text pairs. 
See the supplementary material for the results under SGCls and SGDet.

Table~\ref{tab:predcls} shows an improvement in our approach over rule-based approaches, especially on \textit{Hard}.
That was because the speaker in \textit{Hard} is not in the same frame as the text, while ``frame distance'' gives preference to the character in the same frame. Therefore, other characters are in the frame to which the text belongs, ``frame distance'' is 100\% likely to be wrong.
Besides, the model using reading order information performed better than that without it.

Additionally, we directly tested our model on eBDtheque. 
Our model achieved 77.45\% accuracy under PredCls, slightly below ``frame distance''.
This is mainly due to style variations in old European and American comics, where the speaker and text are often in the same frame. 
More details are available in the supplementary material. 

\subsection{Qualitative Results}
\begin{figure*}[t]
    \centering
    \begin{minipage}[c]{0.34\linewidth}
        \centering
        \includegraphics[width=\linewidth]{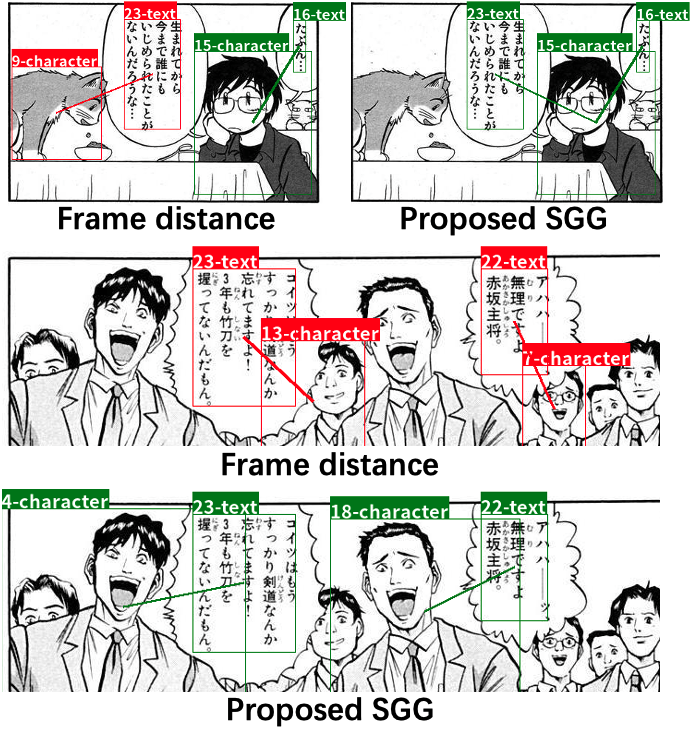}
        \subcaption{Predictions on \textit{Easy}.}
    \end{minipage}
    \begin{minipage}[c]{0.65\linewidth}
        \centering
        \includegraphics[width=0.86\linewidth]{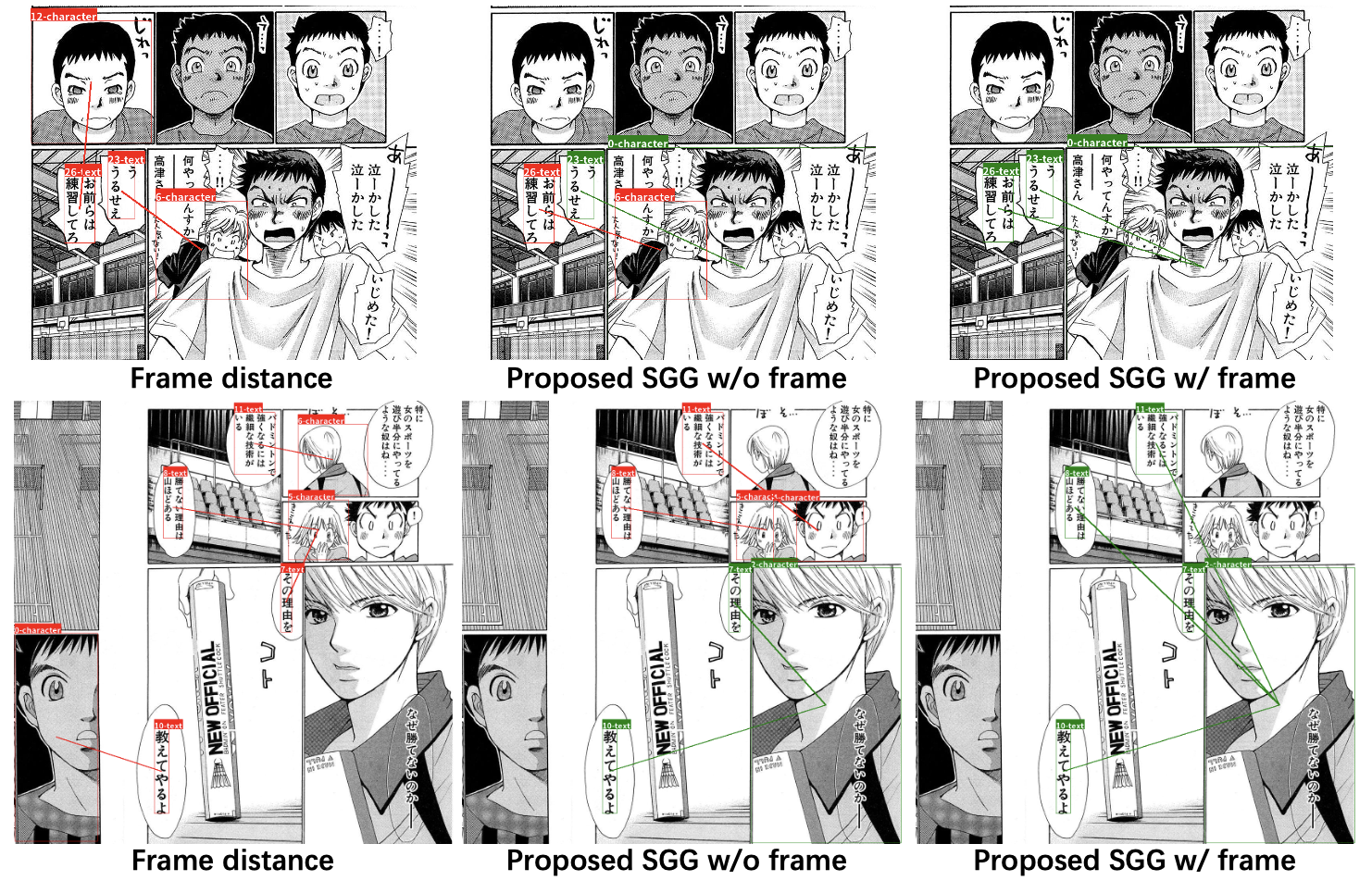}
        \subcaption{Predictions on \textit{Hard}.}
    \end{minipage}
    \caption{Examples of predictions made by rule-based method and proposed method. Courtesy of Gasan, Saijo Shinji, Saki Kaori. The green lines represent correct predictions, while the red lines represent incorrect predictions.}
    \label{fig:vis_success}
\end{figure*}

To better understand the generated scene graphs, we visualized the predictions.
Fig.~\ref{fig:vis_success} shows Recall@(\#text) under PredCls.
The green lines represent correct predictions, and the red lines represent incorrect predictions.
For texts in \textit{Easy}, if the speaker was not the closest character to the text, rule-based methods are certain to make an incorrect prediction, while our model outperformed the ``frame distance'' method regardless of whether frame reading orders were used.
Moreover, the results on \textit{Hard} show that the proposed method of introducing frame orders exhibited a high accuracy in prediction and was able to handle more challenging cases.
We show some additional examples in the supplementary material.

\subsection{Challenges and Future Work}
Although our proposed method demonstrated improvements compared to conventional methods, there were two situations primarily challenging.
(1) The speaker was not in the closest position to the text. Though our model can cope with this situation to some degree, it did not always make correct predictions. 
(2) Texts spoken by different characters appear alternately. Because no visual information can be used for prediction, this may be relatively difficult even for humans. 

The straightforward way to handle these cases would be to introduce Natural Language Processing (NLP) models. 
However, this research aimed to show to what extent speaker detection can be achieved only with visual information. The results of the present study provide a baseline needed to incorporate NLP in future research.


\section{Conclusion}

In this study, we developed Manga109Dialog, a large-scale dialogue dataset that significantly advances the field of comics analysis. 
We pioneered a novel approach by applying SGG models to the task of comics speaker detection, establishing a benchmark for deep learning-based methods where none existed before.
The results not only demonstrate the reliability of Manga109Dialog but also show the superior efficacy of our scene-graph-based approach. 
Our work paves the way for future research, particularly in exploring the effective combination of SGG models and other techniques such as NLP, offering new insights in this field.

\bibliographystyle{IEEEbib}
\bibliography{main}

\clearpage
\onecolumn
\newcommand\beginsupplement{%
        \setcounter{table}{0}
        \renewcommand{\thetable}{\Alph{table}}%
        \setcounter{figure}{0}
        \renewcommand{\thefigure}{\Alph{figure}}%
        \setcounter{equation}{0}
        \renewcommand{\theequation}{\alph{equation}}%
     }
\beginsupplement
\appendix

\section{Dataset Construction Details}
This section shows the construction details of Manga109Dialog.

\subsection{Annotation Rules}
In this section, we use examples to show more details of annotation rules stated in Section 3.1.

\begin{figure}[h]
    \centering
    \begin{minipage}[c]{0.46\linewidth}
        \centering
        \includegraphics[width=0.8\linewidth]{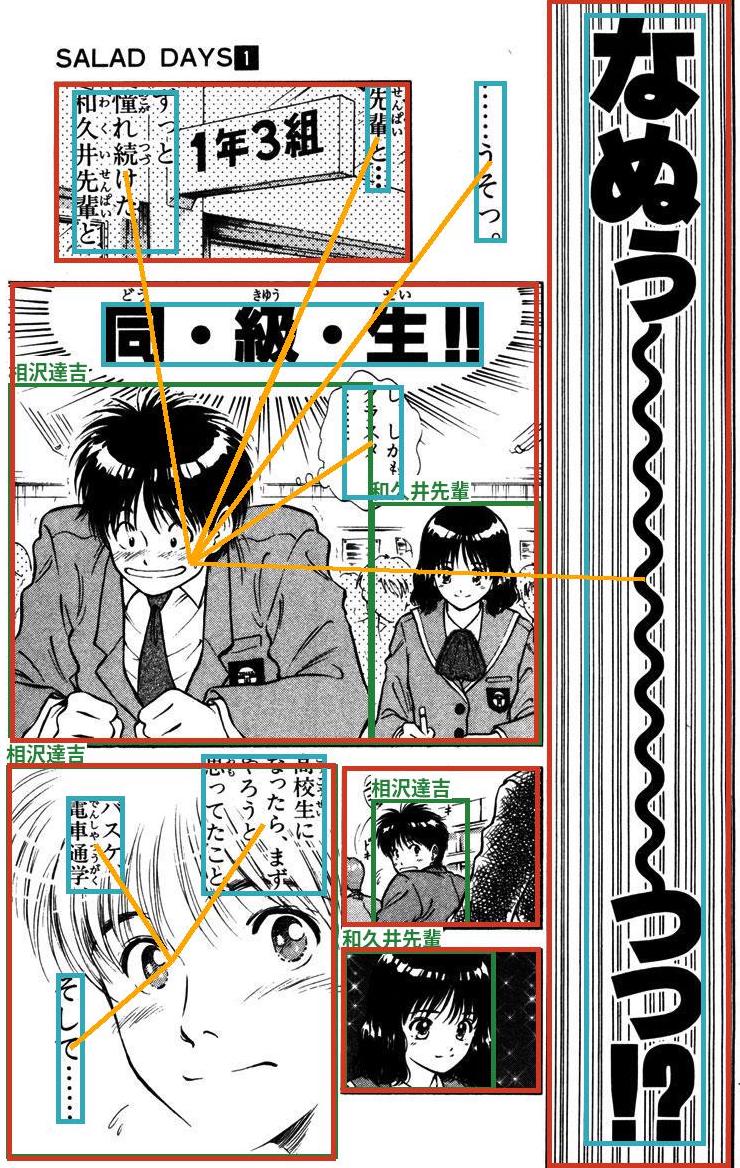}
        \subcaption{If the speaker is not in the same frame, it is determined by the reading order.}
    \end{minipage}
    \quad
    \begin{minipage}[c]{0.48\linewidth}
        \centering
        \includegraphics[width=0.83\linewidth]{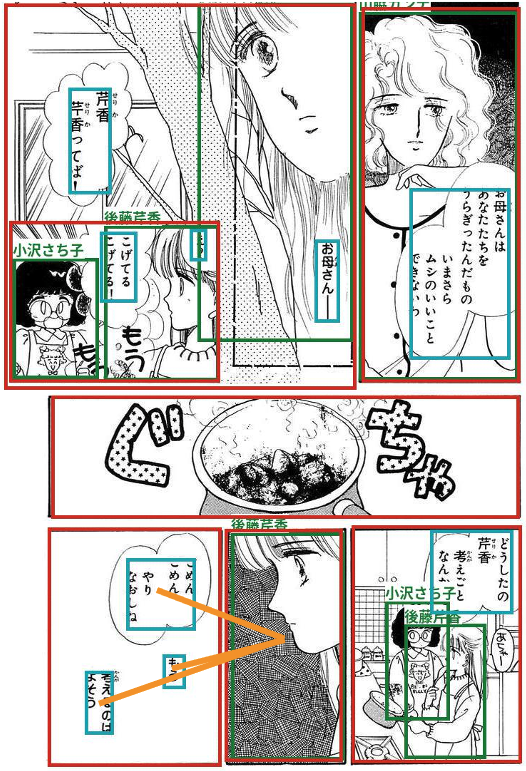}
        \subcaption{If the texts are in the last frame of the page, they are linked to the character in the second-to-last frame.}
    \end{minipage}
    \caption{Examples where the speaker is not in the same frame as the text. Courtesy of Inokuma Shinobu, Kurita Riku.}
    \label{fig:rule1}
\end{figure}

First, let us introduce the basic rules.
Although the images in Manga109 are double-page spreads, from the semantic point of view, we take every single page as a unit.
If the number of the character who is saying the text is only one on the page, link the bounding box of the character to the text regardless of the position.
If the number of the character who is saying the text is more than one on the page, the bounding box of the character in the same frame as the text is linked in priority.
If the character is not in the same frame, it is determined by the reading order (from top to bottom, right to left).
Take the rightmost text of Fig.~\ref{fig:rule1} (a) as an example. Since there is no character in the frame with it, it is linked to the boy who appears in the middle frame according to the reading order.

However, there exist exceptions. For texts in the last frame of the page, since we only consider the single page, we link the character in the second-to-last frame to the text instead of the character in the next page. One example is shown in Fig.~\ref{fig:rule1} (b).

\begin{figure}[h]
    \centering
    \begin{minipage}[c]{0.52\linewidth}
        \centering
        \includegraphics[width=0.9\linewidth]{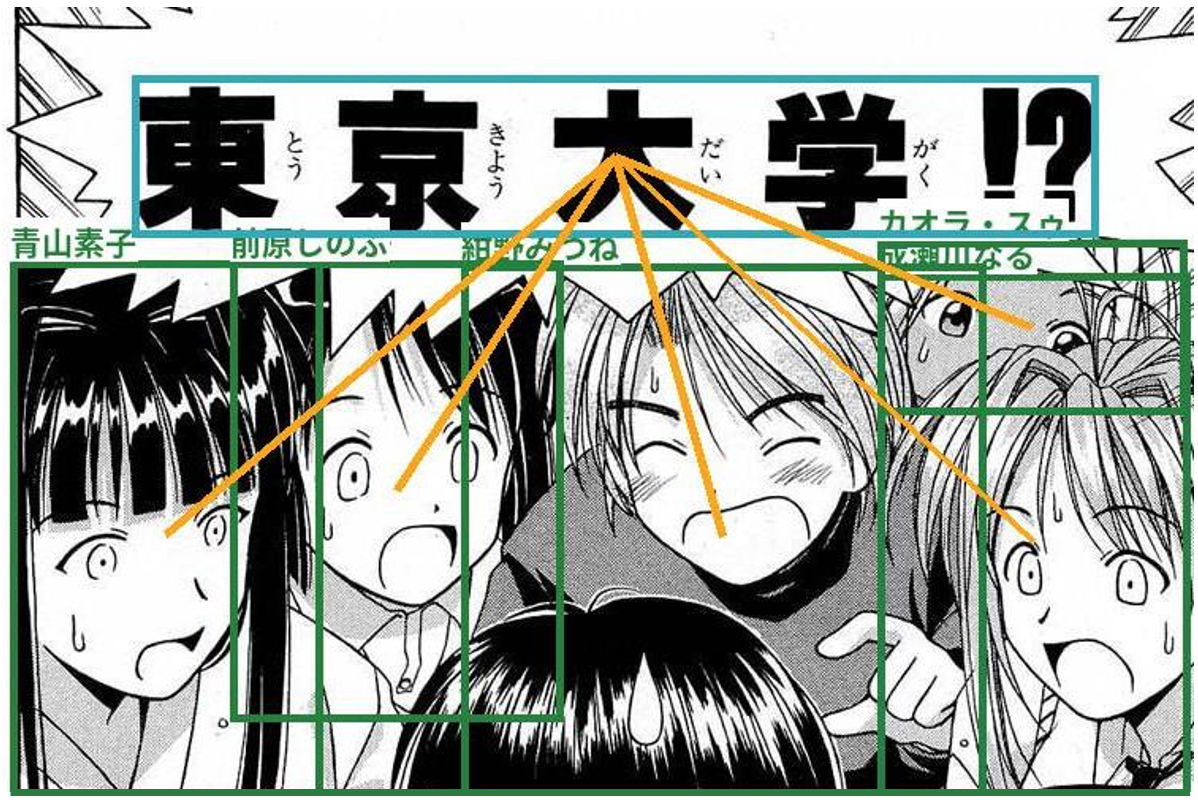}
        \subcaption{An example of multiple speakers for a single text.}
    \end{minipage}
    \begin{minipage}[c]{0.45\linewidth}
        \centering
        \includegraphics[width=\linewidth]{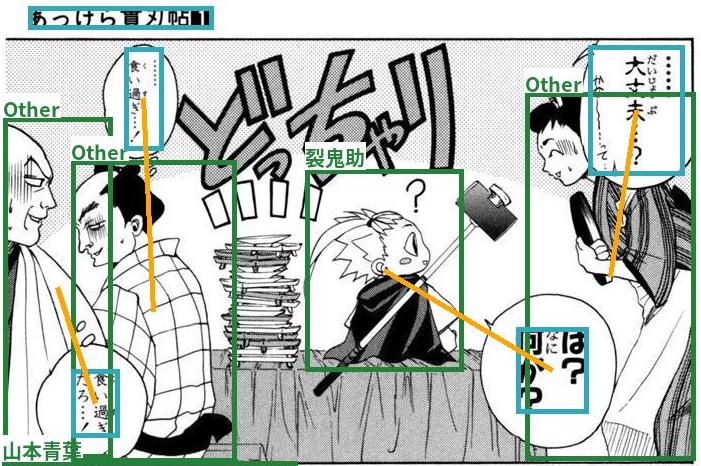}
        \subcaption{An example of texts said by ``Others''.}
    \end{minipage}
    \caption{Some special cases. Courtesy of Akamatsu Ken, Kobayashi Yuki.}
    \label{fig:rule2}
\end{figure}

Next, let us consider some special cases.
Take Fig.~\ref{fig:rule2} (a) as an example. When there is more than one speaker for a single text, we link the text to the bounding boxes of all speakers.

Besides, for texts said by ``Others'', if we can specify the bounding box of the speaker (such as Fig.~\ref{fig:rule2} (b)), it is annotated; otherwise, it is excluded.

\begin{figure}[h]
    \centering
    \begin{minipage}[c]{0.62\linewidth}
        \centering
        \includegraphics[width=0.9\linewidth]{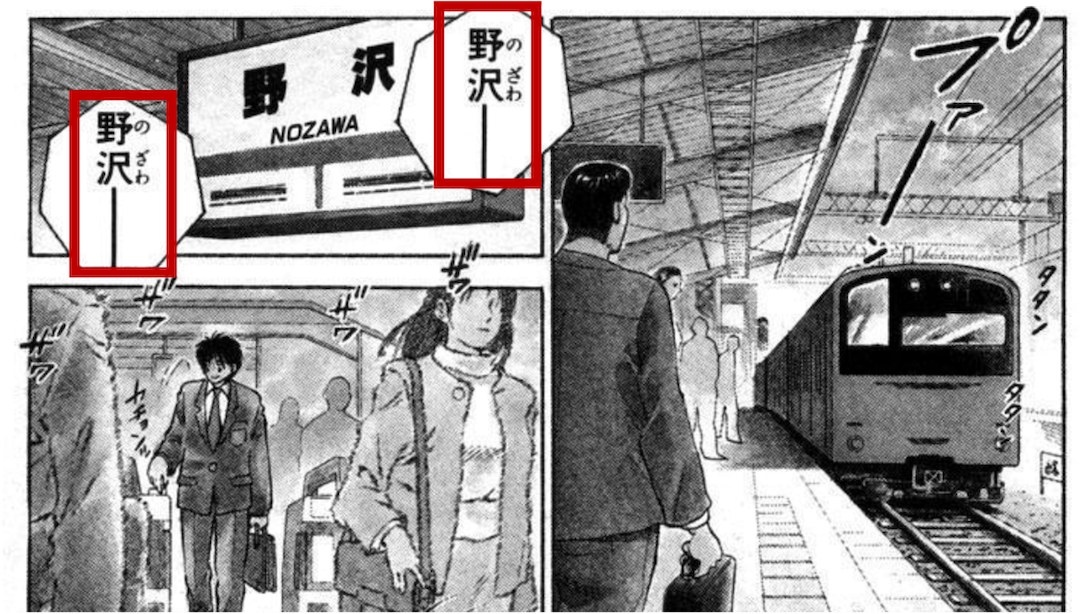}
        \subcaption{Texts with unknown speakers are not our annotation targets.}
    \end{minipage}
    \quad
    \begin{minipage}[c]{0.35\linewidth}
        \centering
        \includegraphics[width=0.95\linewidth]{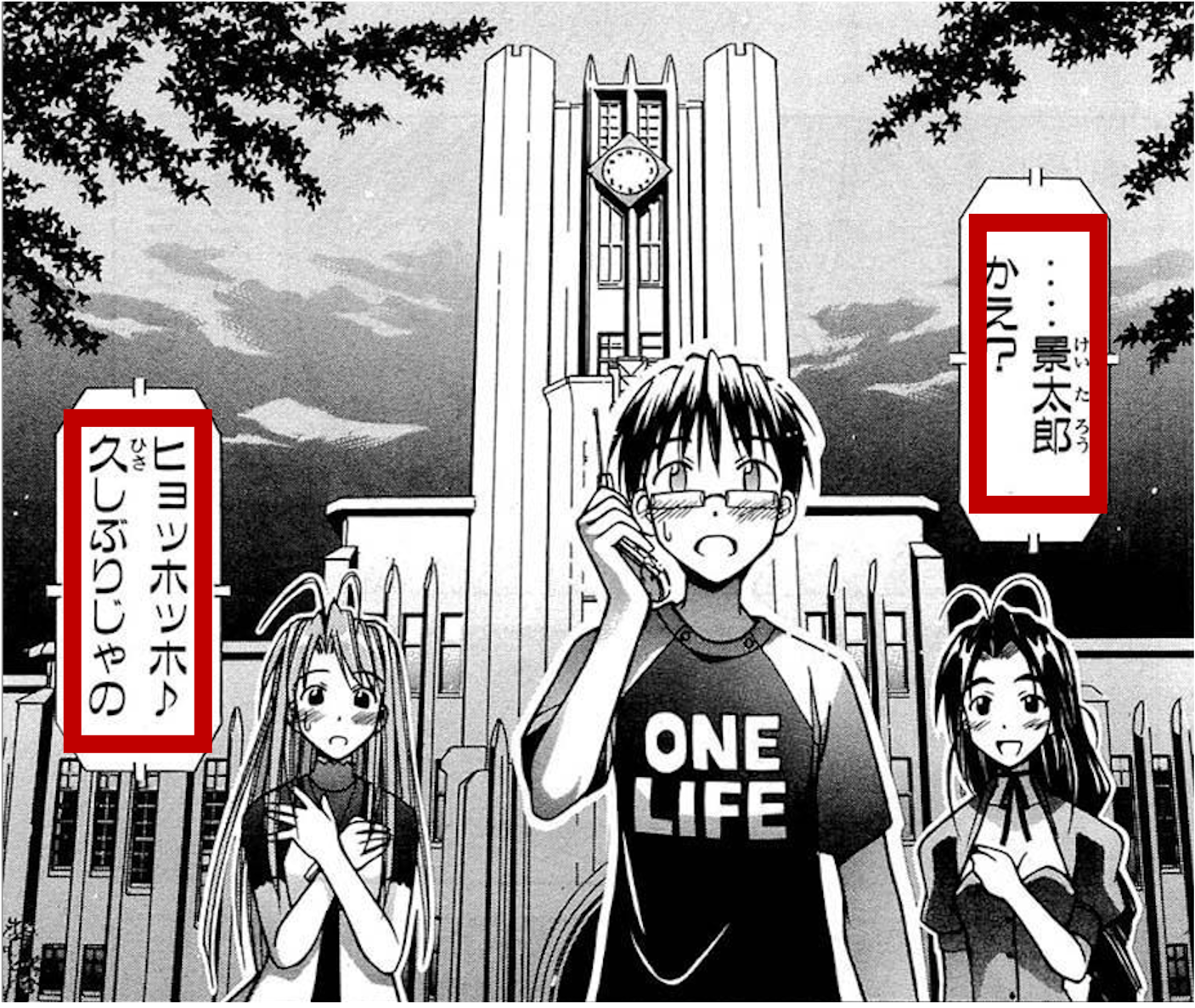}
        \subcaption{When the speaker is not on the page, the text is not annotated.}
    \end{minipage}
    \caption{Texts not in the scope of annotation. Courtesy of Inokuma Shinobu, Akamatsu Ken.}
    \label{fig:rule3}
\end{figure}

Although we make annotations on Manga109, not all texts are our annotation targets.
For example, titles and descriptions are not annotated. 
For texts belonging to ``Narration'' or with unknown speakers, like those in Fig.~\ref{fig:rule3} (a), we do not annotate them, either.
Furthermore, when the speaker is not on the page, such as texts from phone conversations (Fig.~\ref{fig:rule3} (b)) or letters, the text is not in the scope of annotation.

\subsection{Annotation Details}
\begin{figure}[h]
    \centering
    \includegraphics[width=0.75\linewidth]{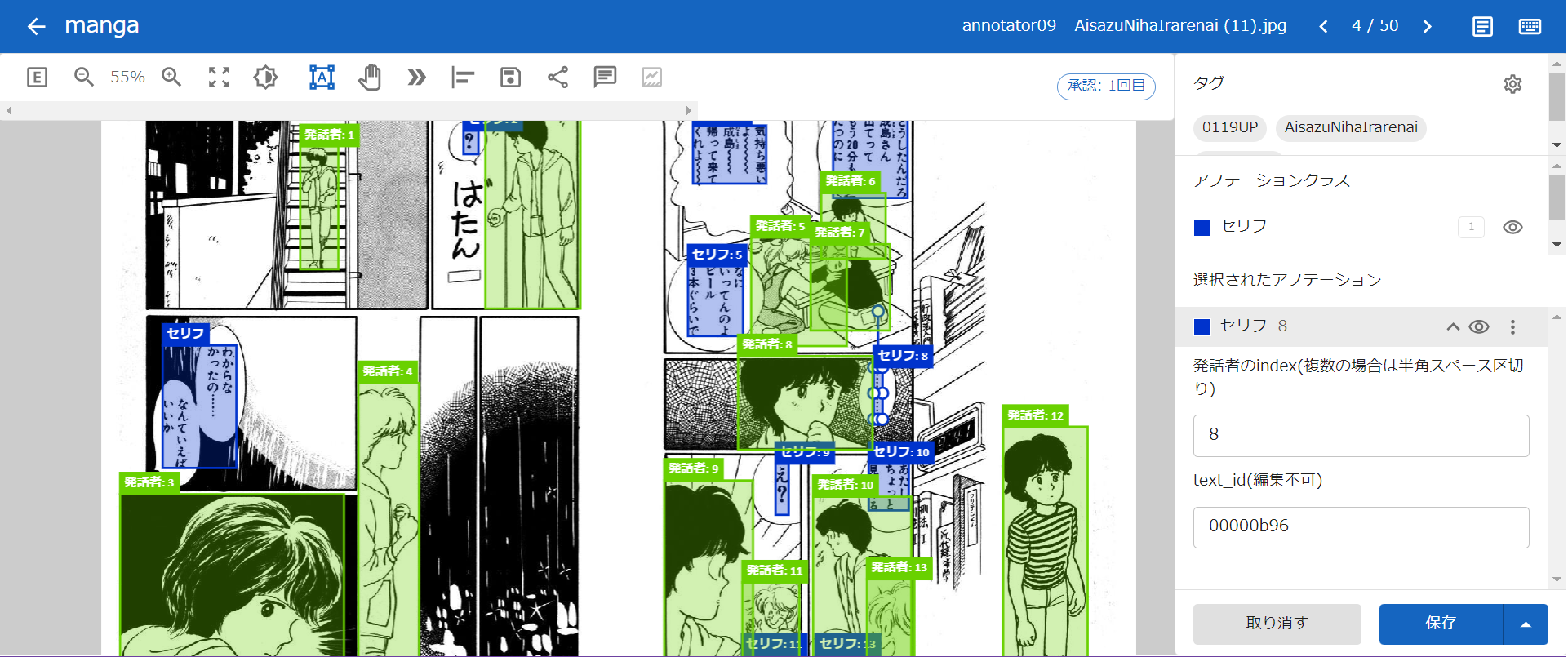}
    \caption{The interface of the annotation tool. Courtesy of Yoshi Masako.}
    \label{fig:anno}
\end{figure}

\begin{figure}[h]
    \centering
    \includegraphics[width=0.75\linewidth]{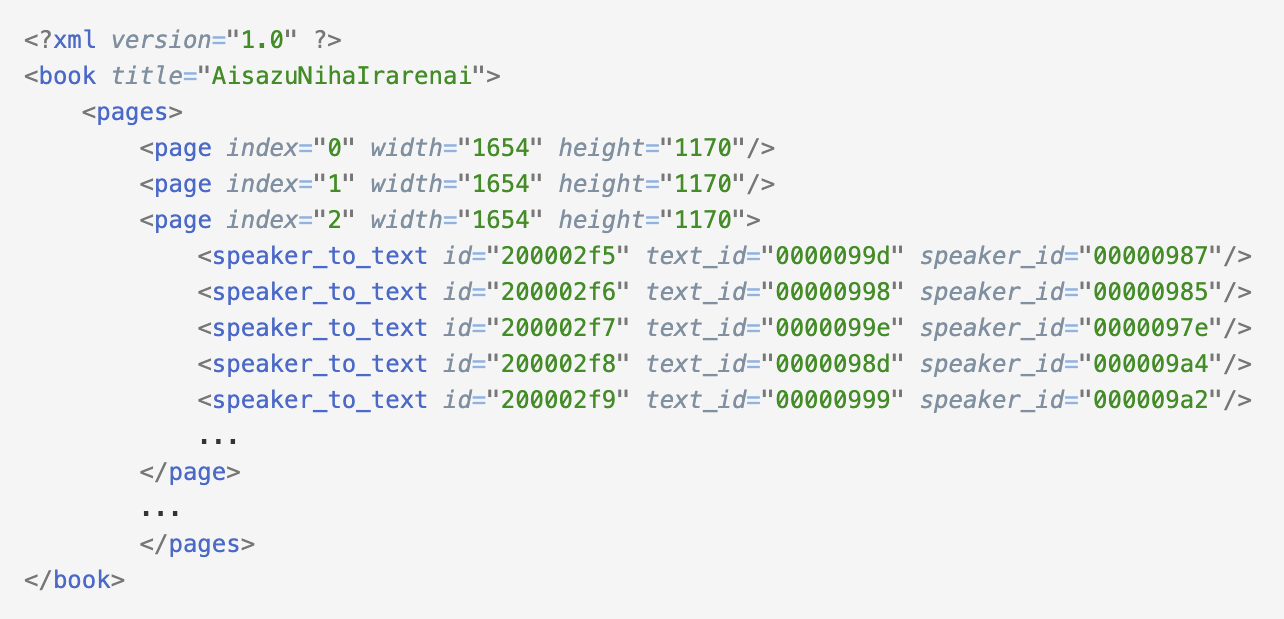}
    \caption{Format of the saved annotations.}
    \label{fig:xml}
\end{figure}

We outsourced the annotation of Manga109Dialog to a company with expert annotators, consisting of 8 operators and 3 inspectors.
Operators used the annotation tool shown in Fig.~\ref{fig:anno} to link the speaker to the text. Then inspectors checked the annotations for omissions and mistakes.
Creating the annotations took approximately three months, and work hours (except for inspections) exceeded 627 hours.
The annotations for each comic are saved in an xml file, an example is shown in Fig.~\ref{fig:xml}.

\subsection{Visualization Tool}
\begin{figure}[h]
    \centering
    \includegraphics[width=0.8\linewidth]{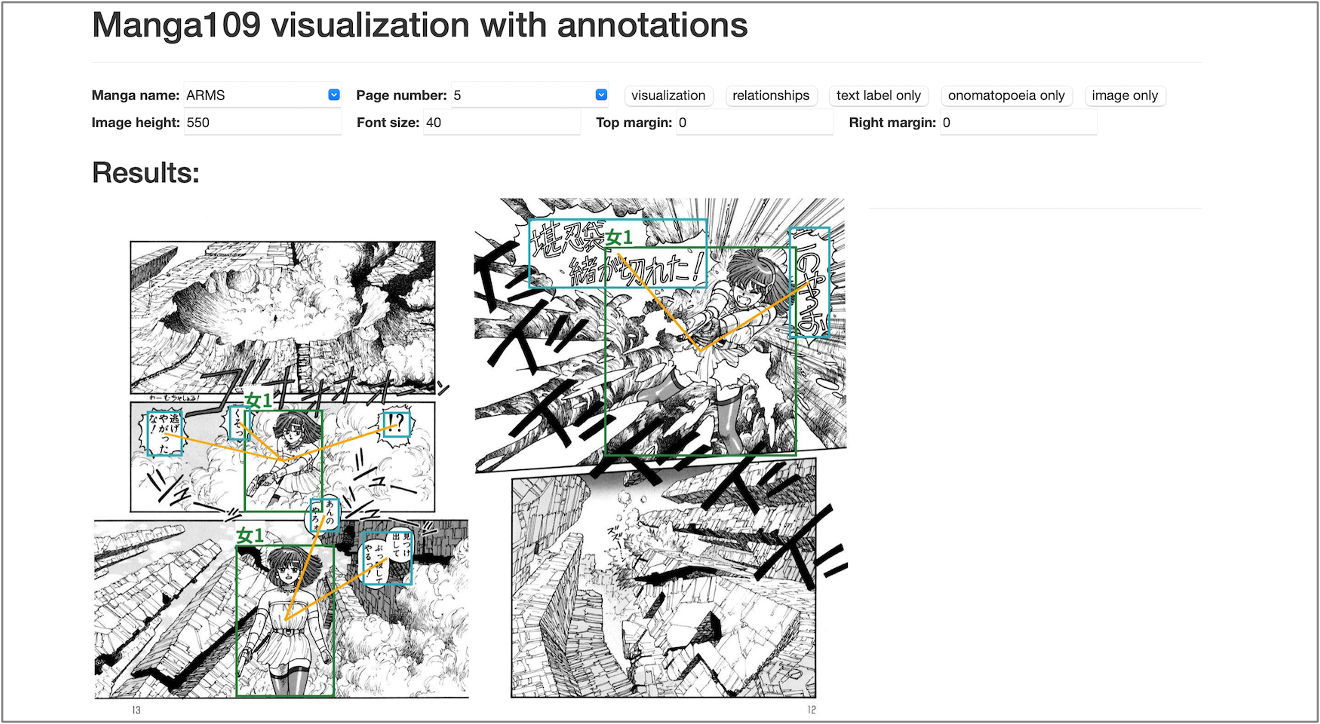}
    \caption{The interface of our visualization tool. Courtesy of Kato Masaki.}
    \label{fig:vis}
\end{figure}

After the annotation company completed the task, we confirmed and corrected their annotations.
To check the annotations in a visual way, we designed a visualization tool. The interface is shown in Fig.~\ref{fig:vis}.

When pressing the ``relationship'' button, the visualization tool can show the regions of characters and texts with different colored rectangular boxes. Additionally, it can show characters' names and speaker-to-text pairs. 
As shown in Fig.~\ref{fig:vis}, each orange line connects the center coordinates of the speaker and the text.
When a text has $n$ speakers, there will be $n$ lines.

\clearpage
\section{Experimental Preparation}
This section visually explains the details of our pre-trained object detector, frame detector, and frame order estimator. 

\begin{figure}[H]
    \centering
    \begin{minipage}[c]{0.48\linewidth}
        \centering
        \includegraphics[width=0.8\linewidth]{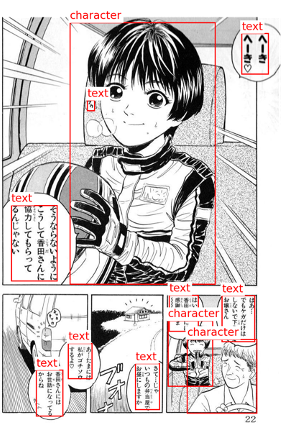}
        \subcaption{Ground truth.}
    \end{minipage}
    \begin{minipage}[c]{0.48\linewidth}
        \centering
        \includegraphics[width=0.8\linewidth]{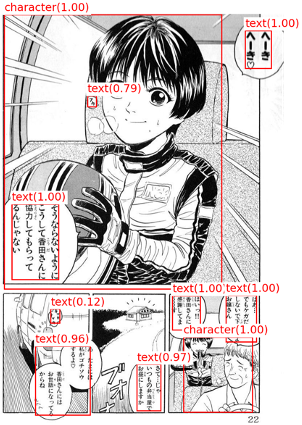}
        \subcaption{Detection results.}
    \end{minipage}
    \caption{An example of detection results. Courtesy of Matsuda Naomasa.}
    \label{fig:det_results}
\end{figure}

\subsection{Object Detector}
We pre-trained our object detector to save run time of speaker prediction.
Following the previous studies in SGG~\cite{tang2020unbiased}, we applied the Faster-RCNN with a ResNeXt-101-FPN backbone.
We then froze the model as the object detector.
We trained our models on a single NVIDIA A100 GPU, with a batch size of 4. 
We used SGD as an optimizer and set the initial learning rate to $4 \times 10^{-3}$.

We set object classes to three: \textit{character}, \textit{text} and \textit{background}.
The mAP of our detector was 86.09\%. The high accuracy may be related to the small number of object categories in our experiments.
An example of detection results is shown in Fig.~\ref{fig:det_results}.
The left side of Fig.~\ref{fig:det_results} shows the visualization of ground truth, and the right side shows the detected bounding boxes and predicted labels.
We can also see from this example that our model has a high accuracy.



\subsection{Frame Detector and Frame Order Estimator}
We pre-trained another Faster-RCNN model as the frame detector, which was trained on a single NVIDIA A100 GPU, with a batch size of 4. 
We used SGD as an optimizer and set the initial learning rate to $4 \times 10^{-3}$.
We set object classes to two: \textit{frame} and \textit{background} and our frame detector achieved an mAP of 96.48\%.

For detected frames, we used the algorithm described in Kovanen and Ikuta's study~\cite{kovanen2015layered,ikuta2023} to presume their reading orders.
Given a set of frame bounding boxes, we first decide whether they are horizontally divisible into two parts. If so, we split them and repeat the first step for each set until we cannot horizontally split them.
Then, we decide whether each set is vertically divisible into two parts. If so, we do so and repeat the first step.
When the set of frames cannot be divided horizontally and vertically, we number the frames according to reading order (from top to bottom, from right to left).

\begin{figure}[t]
    \centering
    \begin{minipage}[c]{0.48\linewidth}
        \centering
        \includegraphics[width=0.95\linewidth]{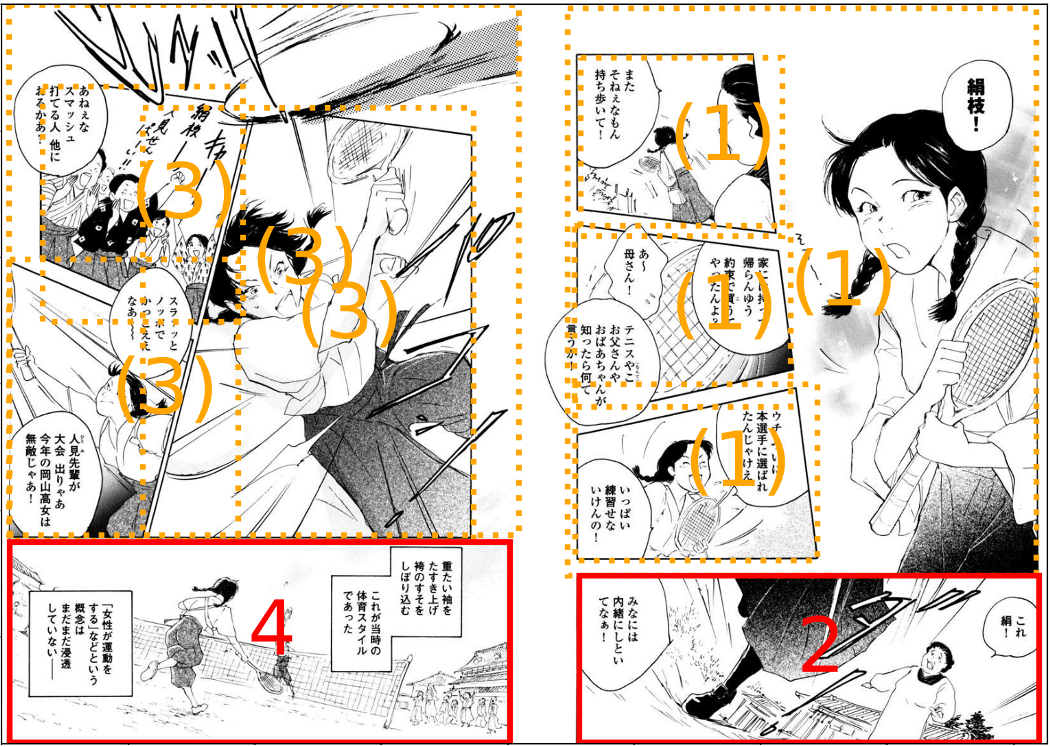}
        \subcaption{Ground truth.}
    \end{minipage}
    \begin{minipage}[c]{0.48\linewidth}
        \centering
        \includegraphics[width=0.95\linewidth]{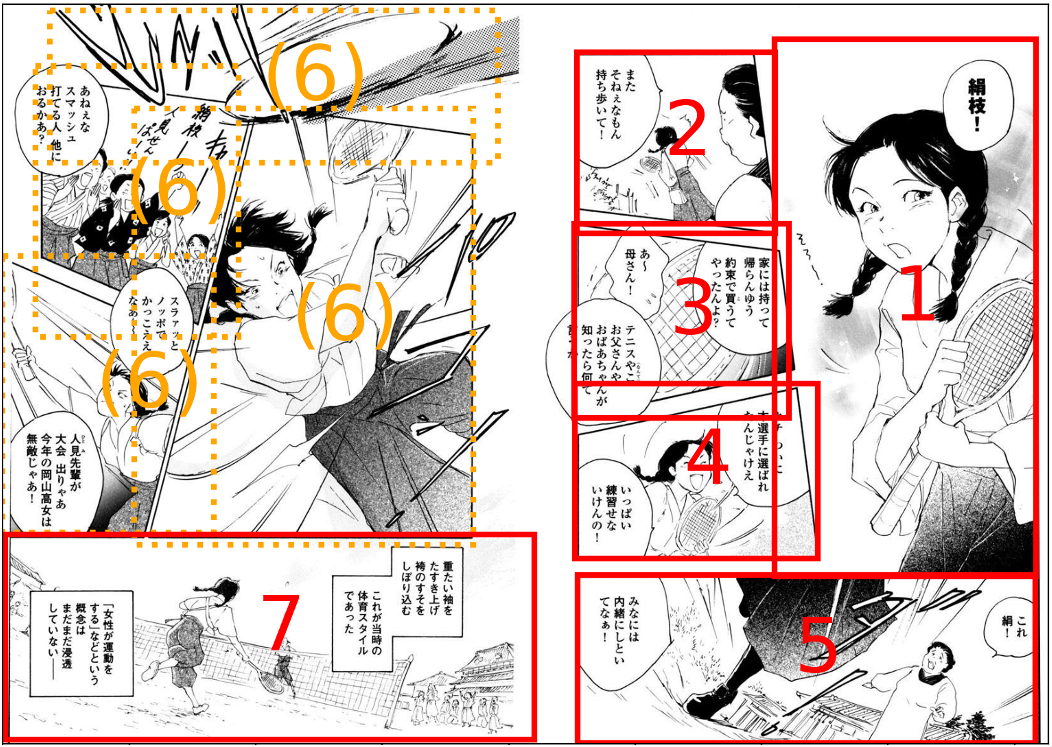}
        \subcaption{Detection results.}
    \end{minipage}
    \caption{Results of frame detection and reading order estimation. Courtesy of Hikochi Sakuya.}
    \label{fig:frame_results}
\end{figure}

An example of the estimation results is shown in Fig.~\ref{fig:frame_results}.
When there is a high degree of overlap in a set of frames, we consider the set inseparable. All frames in this set are given the same order and are shown in orange dotted-line bounding boxes.
For the example in Fig.~\ref{fig:frame_results}, the first frame covers half of the page in ground truth. Since the following three frames are inside the first frame, the first four frames are considered inseparable and share the same reading order.
Therefore, although the detected frame is different from those in ground truth, the reading order is presumed correctly, so there is no negative impact on our experiments of speaker prediction.

\section{Rule-Based Baseline}
\label{sec:rule-based}
Here, we recap simple rule-based approaches using our notations in Section 4.2.

The simplest baseline is ``shortest distance''. That is, the score $r_{i \to j}$ is defined as follows.
\begin{equation}
r_{i \to j} = 
\begin{cases}
1 & \mathrm{if}~\mathbf{b}_i~\mathrm{is~closest~to}~\mathbf{b}_j, \\
  & l_i = \mathtt{character},~~\mathrm{and}~~l_j = \mathtt{text}. \\
0 & \mathrm{else}.
\end{cases}
\label{eq:shortest_dis}
\end{equation}
Here, we assume that $\{\mathbf{b}_i\}_{i=1}^N$ is given by $f_\mathrm{detect}$. This equation means that the text is spoken by the closest character. This is simple and can be computed rapidly, but cannot handle complex cases.

A slightly modified version is described as ``frame distance'', which is defined as given below.
\begin{equation}
r_{i \to j} = 
\begin{cases}
1 & \mathrm{if}~\mathbf{b}_i~\mathrm{is~closest~to}~\mathbf{b}_j, \\
  & l_i = \mathtt{character},~~l_j = \mathtt{text},~~\mathrm{and}\\
  & \mathrm{both}~\mathbf{b}_i, \mathbf{b}_j~\mathrm{belong~to~the~same}~\mathbf{p} \\
\mathrm{Eq.~\ref{eq:shortest_dis}} & \mathrm{else}.
\end{cases}
\label{eq:frame_dis}
\end{equation}
Here, we prioritize characters in the same frame as the text. If there are no characters in the frame, we make predictions using the same rule as ``shortest distance''.
In these two approaches, the amount of $r$ is the same as that of $\mathtt{text}$. Therefore, there is no need to select the top $K$ predictions.

\section{Ablation Studies}
To evaluate our proposed method, we conducted ablation studies and ran our model under the following four patterns.
\begin{itemize}[leftmargin=*]
    \setlength{\itemsep}{-0.1ex}
    \item Rule-based: methods discribed in Section~\ref{sec:rule-based}. In ``frame distance'', the frames are obtained from our frame detector.
    \item Only w/ union feature: a deep learning-based simple baseline that calculates relationship scores only using the union feature $\textbf{f}_{i, j}$ obtained in Eq. 4, without using the fusion module $f_\mathrm{fuse}$ in Eq. 2.
    \item Proposed w/o frame: predict speakers without frame information. In this case, the process is the same as the original SGG process.
    \item Proposed w/ detected frame: use the frame bounding boxes detected by Faster-RCNN. This scenario is a practical use case. We detect both the frame positions and the reading orders.
    \item Proposed w/ given frame: use the frame bounding boxes from Manga109's annotations. This means that ground truth bounding boxes of frames are available. We then estimate the frame orders.
\end{itemize}

Recall@(\#text) under three SGG tasks is shown in Table~\ref{tab:ablation_predcls},~\ref{tab:ablation_sgcls},~\ref{tab:ablation_sgdet}.

From these tables, we can observe a significant improvement for the model with frame information over the model without it. 
Moreover, the model with given frames performs better than the model with detected frames overall, but ``Proposed w/ detected frame'' overperforms ``Proposed w/ given frame'' on \textit{Hard}.
It may be because the speaker and the text are not in the same frame in data on \textit{Hard}, while our frame detector divided frames into smaller ones, making the estimated reading order more accurate.

\section{Experiments with Different Styles of Comics}
To see the behavior of our model on different styles of comics or languages, we tested our model on eBDtheque~\cite{guerin2013ebdtheque}, which comprises 100 images from French, American, and Japanese comics. Results under PredCls is shown in Table~\ref{tab:ebdtheque}.
From the table, it may be observed that although our model was trained on Manga109, it also performed well on eBDtheque.
Our method outperformed ``shortest distance'' but did not perform as well as ``frame distance''.
This was mainly because, in contrast to Japanese manga, the speaker and the text are almost always in the same frame in relatively old European and American comics.
This experiment highlights the variations in comic styles across different countries and outlines directions for future research.

\begin{table}[h]
    \centering
    \begin{minipage}[c]{0.48\linewidth}
    \centering
    \caption{Recall@(\#text) for PredCls.}
    \begin{tabular}{@{}llll@{}} \toprule
              & \multicolumn{3}{c}{Manga109Dialog} \\ \cmidrule(l){2-4}
        Method & \textit{Easy} & \textit{Hard} & \textit{Total} \\ \midrule
        Rule-based               & & & \\
        ~~~~Short distance       & 71.43 & 22.72 & 63.41 \\
        ~~~~Frame distance       & 81.55 & 22.06 & 71.53 \\
        Deep learning-based             & & & \\
        ~~~~Only w/ union feature   & 83.20 & 28.50 & 73.99 \\
        ~~~~Proposed w/o frame   & 84.73 & 29.55 & 75.48 \\   
        ~~~~Proposed w/ detected frame      & 84.77 & \textbf{30.65} & 75.69 \\
        ~~~~Proposed w/ given frame         & \textbf{85.24} & 29.50 & \textbf{75.89} \\ \bottomrule    
    \end{tabular}
    \label{tab:ablation_predcls}
    \end{minipage}
    \begin{minipage}[c]{0.48\linewidth}
    \centering
    \caption{Recall@(\#text) for SGCls.}
    \begin{tabular}{@{}llll@{}} \toprule
              & \multicolumn{3}{c}{Manga109Dialog} \\ \cmidrule(l){2-4}
        Method & \textit{Easy} & \textit{Hard} & \textit{Total} \\ \midrule
        Rule-based               & & & \\
        ~~~~Short distance       & 71.06 & 23.71 & 63.09 \\
        ~~~~Frame distance       & 80.81 & 20.61 & 70.68 \\
        Deep learning-based             & & & \\
        ~~~~Only w/ union feature   & 82.74 & 28.41 & 73.59 \\
        ~~~~Proposed w/o frame      & 84.42 & 29.12 & 74.92 \\   
        ~~~~Proposed w/ detected frame      & 84.51 & \textbf{29.77} & 75.32 \\
        ~~~~Proposed w/ given frame         & \textbf{84.97} & 29.23 & \textbf{75.62} \\ \bottomrule   
    \end{tabular}
    \label{tab:ablation_sgcls}
    \end{minipage}
\end{table}

\begin{table}[h]
    \centering
    \begin{minipage}[c]{0.48\linewidth}
    \centering
    \caption{Recall@(\#text) for SGDet.}
    \begin{tabular}{@{}llll@{}} \toprule
              & \multicolumn{3}{c}{Manga109Dialog} \\ \cmidrule(l){2-4}
        Method & \textit{Easy} & \textit{Hard} & \textit{Total} \\ \midrule
        Rule-based               & & & \\
        ~~~~Short distance       & 53.11 & 14.85 & 46.67 \\
        ~~~~Frame distance       & 64.49 & 13.67 & 55.93 \\
        Deep learning-based             & & & \\
        ~~~~Only w/ union feature   & 67.48 & 21.31 & 59.71 \\
        ~~~~Proposed w/o frame      & 67.84 & 21.43 & 60.04 \\  
        ~~~~Proposed w/ detected frame      & 67.92 & \textbf{24.40} & 60.54 \\
        ~~~~Proposed w/ given frame         & \textbf{68.21} & 22.96 & \textbf{60.60} \\ \bottomrule    
    \end{tabular}
    \label{tab:ablation_sgdet}
    \end{minipage}
    \begin{minipage}[c]{0.48\linewidth}
    \centering
    \caption{Experimental results on eBDtheque.}
    \begin{tabular}{@{}lcc@{}}
    \toprule
    Method & eBDtheque \\
    \midrule
    Rule-based               & \\
    ~~~~Short distance       & 77.34 \\
    ~~~~Frame distance       & \textbf{80.27} \\
    Deep learning-based      & \\
    ~~~~Proposed (SGG) w/o frame & 75.20 \\
    ~~~~Proposed (SGG) w/ frame & 77.45 \\
    \bottomrule
    \end{tabular}
    \label{tab:ebdtheque}
    \end{minipage}
\end{table}

\clearpage
\section{Experiments with Other SGG Models}
This section explains why we chose MOTIFS as the SGG model for our proposed pipeline.
In addition to MOTIFS, we ran experiments using other SGG models.
We changed the LSTMs in the fusion module $f_\mathrm{fuse}$ to TreeLSTMs of VCTree~\cite{tai2015improved,tang2019learning} and Transformer~\cite{vaswani2017attention,tang2020unbiased}, respectively.
Experimental results are shown in Table~\ref{tab:diff_sgg}.

From the table, we can see that the methods using the SGG models all perform better than the rule-based method and DL simple baseline. 
However, there is not a large gap between the results of the different SGG models. 
It is probably because the relationship is only one class in our experiments, and it is difficult for models such as LSTMs to exert their advantages.
\begin{table}[h]
    \centering
    \caption{Results using different SGG models.}
    \begin{tabular}{@{}llll@{}} \toprule
    & \multicolumn{3}{c}{Manga109Dialog} \\ \cmidrule(l){2-4}
    Method & \textit{Easy} & \textit{Hard} & \textit{Total} \\ \midrule
    Frame distance      & 81.55 & 22.06 & 71.53 \\
    DL simple baseline  & 83.20 & 28.50 & 73.99 \\
    Proposed SGG        & & & \\
    ~~~~MOTIFS          & \textbf{84.73} & \textbf{29.55} & \textbf{75.48} \\   
    ~~~~VCTree          & 83.30 & 28.85 & 74.16 \\
    ~~~~Transformer     & 83.76 & 28.63 & 74.51 \\ \bottomrule   
    \end{tabular}
    \label{tab:diff_sgg}
\end{table}

\section{Qualitative Results}
\begin{figure}[h]
  \centering
  \includegraphics[width=0.98\linewidth]{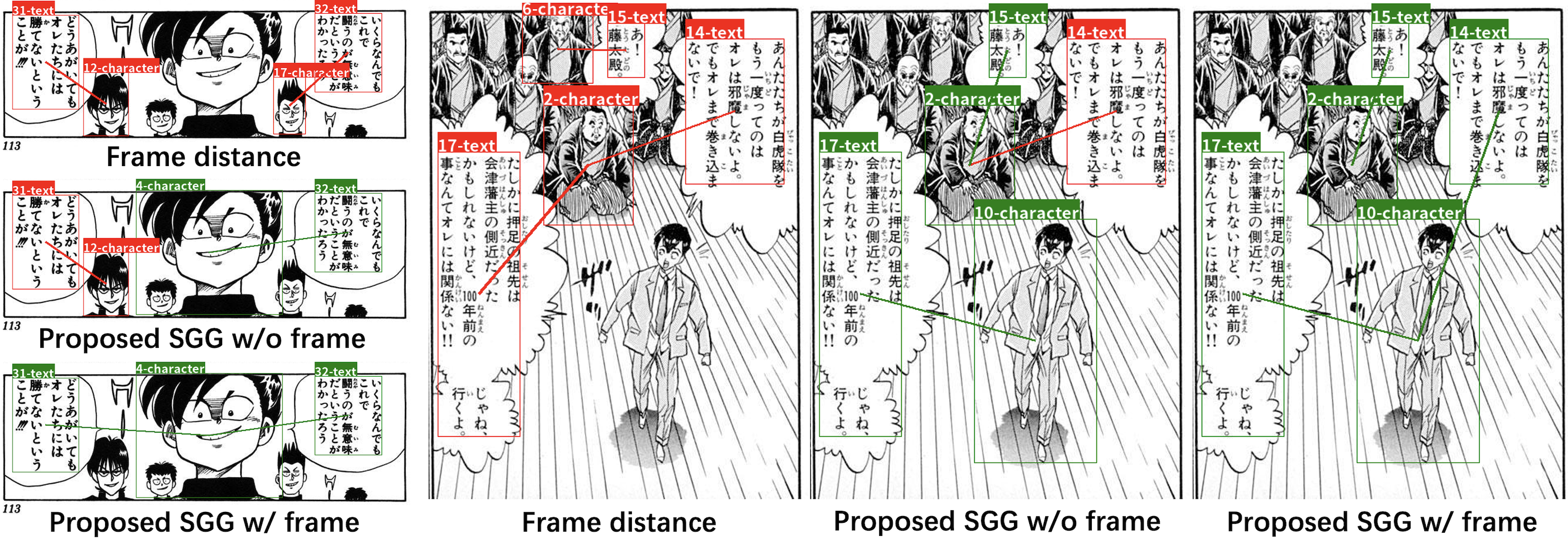}
  \caption{Predictions on \textit{Easy}. Courtesy of Yabuno Tenya, Watanabe Tatsuya, Saijo Shinji. The green lines represent correct predictions, while the red lines represent incorrect predictions.}
  \label{fig:examples1}
\end{figure}

\begin{figure}[t]
  \centering
  \includegraphics[width=0.96\linewidth]{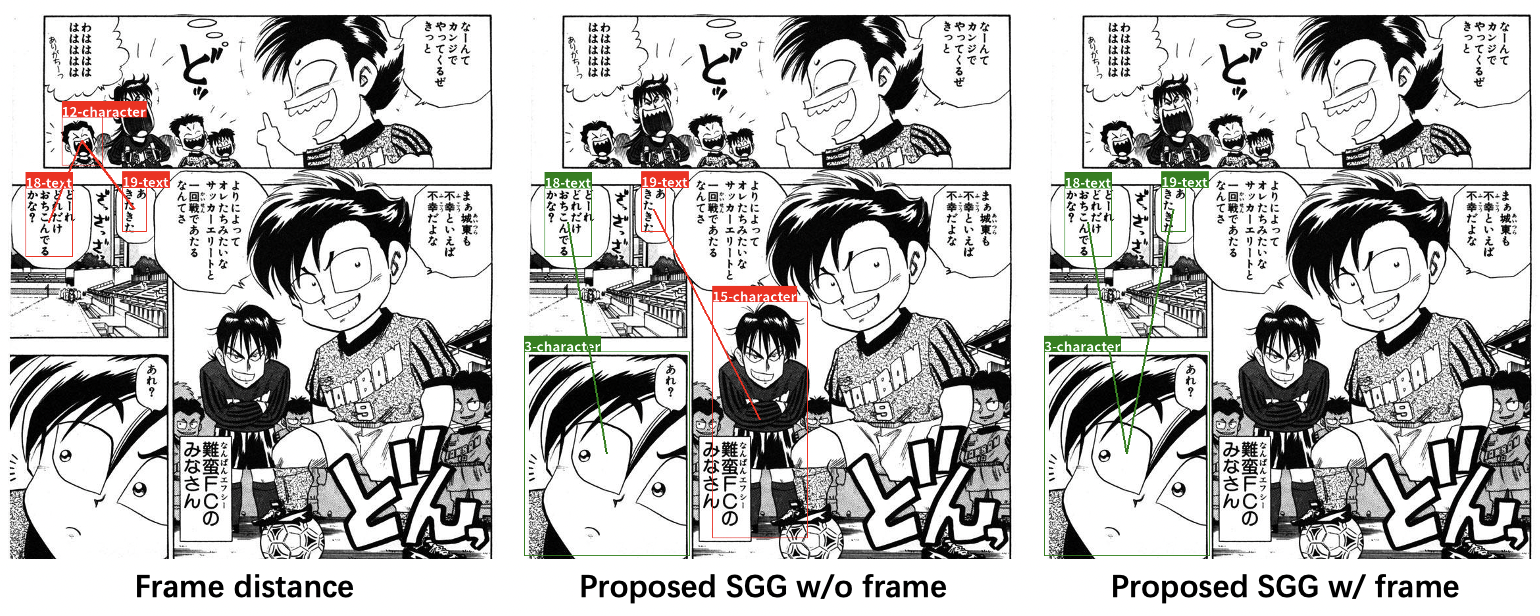}
  \caption{Predictions on \textit{Hard}. Courtesy of Yabuno Tenya, Watanabe Tatsuya.}
  \label{fig:examples2}
\end{figure}

\begin{figure}[t]
  \centering
  \includegraphics[width=0.98\linewidth]{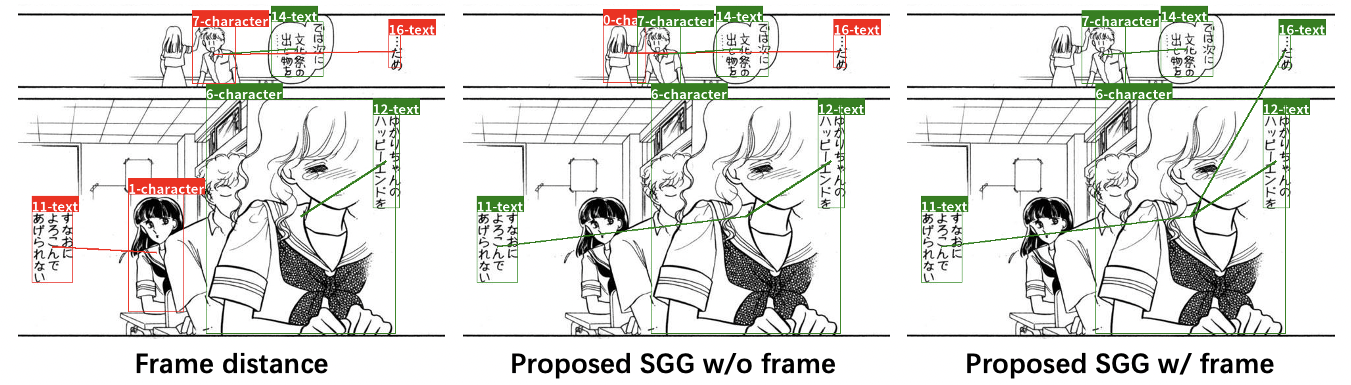}
  \caption{Predictions on \textit{Total}. Courtesy of Tashiro Kimu.}
  \label{fig:examples3}
\end{figure}

\begin{figure}[t]
  \centering
  \begin{minipage}[c]{0.8\linewidth}
      \centering
      \fbox{\includegraphics[width=0.66\linewidth]{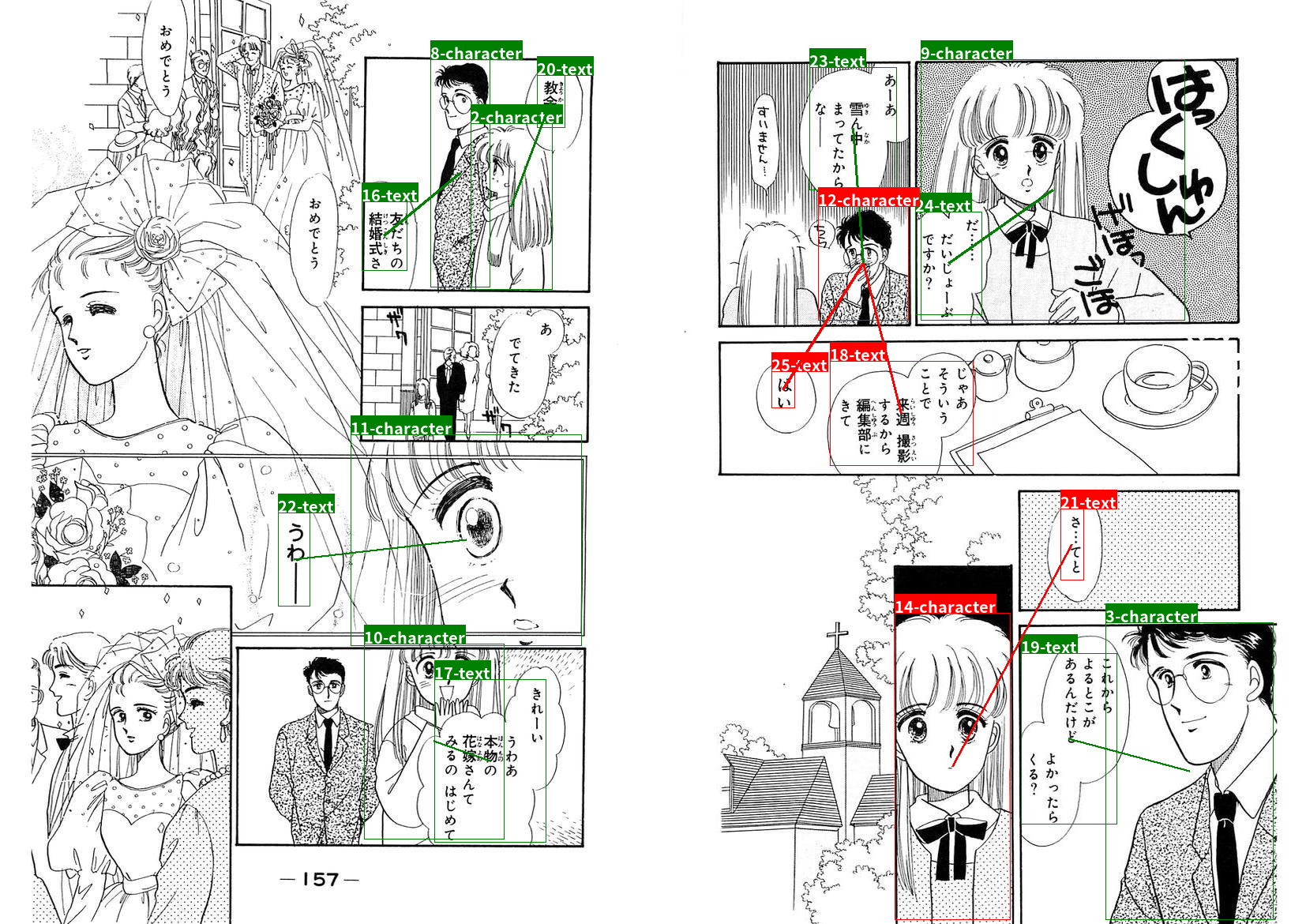}}
      \subcaption{Frame distance.}
  \end{minipage}%
  \vspace{8pt}
  \begin{minipage}[c]{0.8\linewidth}
      \centering
      \fbox{\includegraphics[width=0.66\linewidth]{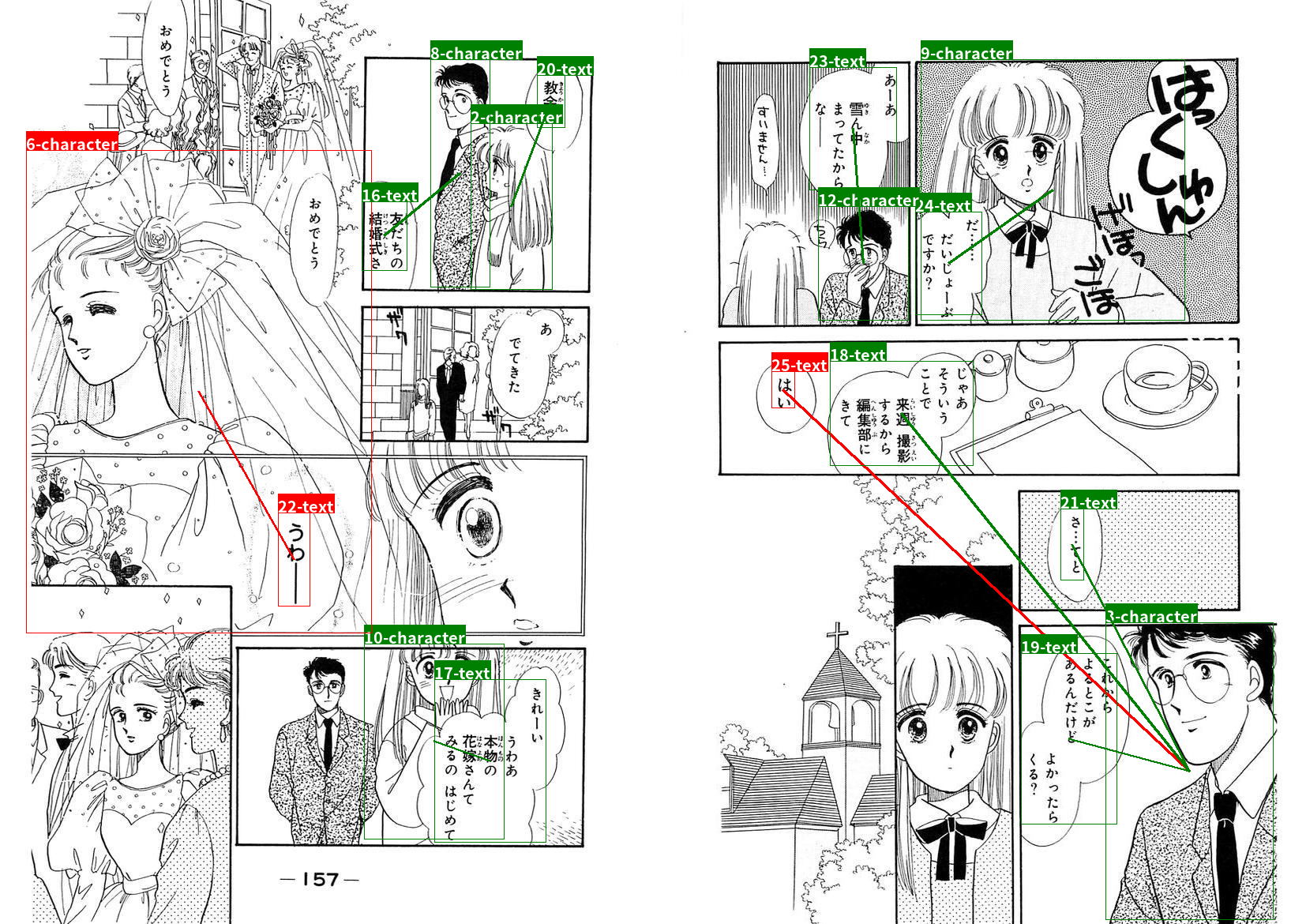}}
      \subcaption{Proposed SGG w/o frame.}
  \end{minipage}%
  \vspace{8pt}
  \begin{minipage}[c]{0.8\linewidth}
      \centering
      \fbox{\includegraphics[width=0.66\linewidth]{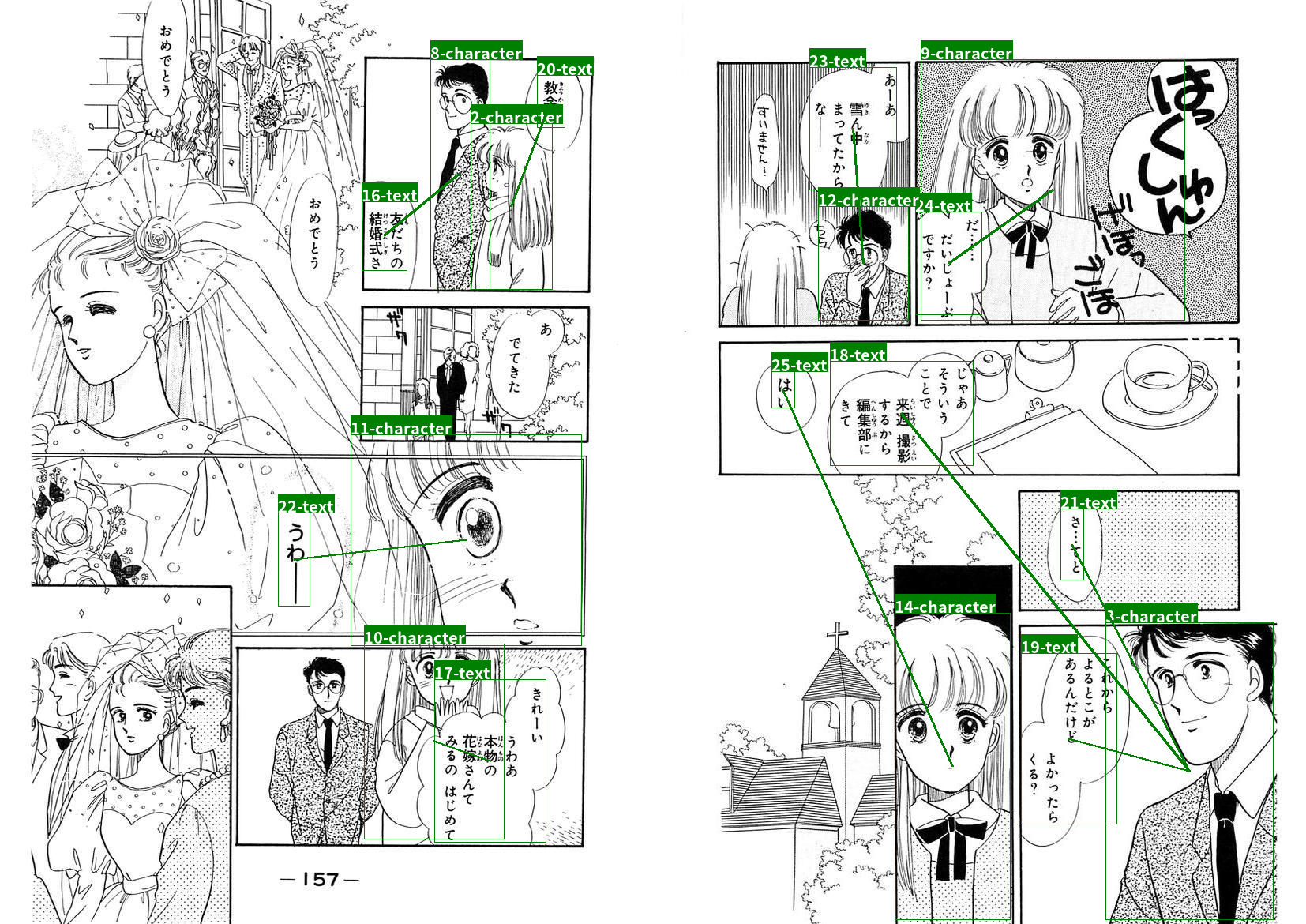}}
      \subcaption{Proposed SGG w/ frame.}
  \end{minipage}
  \caption{Predictions on \textit{Total}. Courtesy of Kurita Riku.}
  \label{fig:examples4}
\end{figure}

In this section, we show more examples of prediction results. 
Speaker predictions on \textit{Easy}, \textit{Hard} and \textit{Total} are given in Fig.~\ref{fig:examples1}, Fig.~\ref{fig:examples2}, Fig.~\ref{fig:examples3} and Fig.~\ref{fig:examples4}, which show Recall@(\#text) in the task of PredCls. The green lines represent correct predictions, while the red lines represent incorrect predictions.

As we can see, our proposed model shows better performance and can deal with more complex cases, especially when there is at least one character in the same frame as the text. 
In this case, if the speaker is not the closest character to the text (like Fig.~\ref{fig:examples1} and the lower part in Fig.~\ref{fig:examples3}) or the speaker is not in the same frame (like the upper part in Fig.~\ref{fig:examples3}), ``frame distance'' will surely make a wrong prediction, while our model can ease this problem to some degree.

\clearpage

\end{document}